\documentclass{article} 
\usepackage{meta/fancyhdr, meta/natbib, meta/iclr2026_conference,times}

\usepackage[hidelinks,
    colorlinks=true,
    linkcolor=bleuhtml!90!white, 
    citecolor=bleuhtml!90!white, 
    filecolor=orange, 
    urlcolor=violet,]   
    {hyperref}
\usepackage[english]{babel}
\usepackage{url}
\usepackage[table, dvipsnames]{xcolor}
\usepackage{xspace}  
\usepackage{amssymb}
\usepackage{graphicx}
\usepackage{booktabs}
\usepackage{subcaption}
\usepackage{wrapfig}
\usepackage{multirow}
\usepackage{amsmath}

\usepackage{tikz}
\usetikzlibrary{positioning}
\usepackage{pgfplots}
\pgfplotsset{compat=1.18}
\usepgfplotslibrary{statistics}

\usepackage[capitalize]{cleveref}
\crefname{figure}{Fig.}{Figs.}
\Crefname{figure}{Figure}{Figures}
\crefname{section}{Sec.}{Secs.}
\Crefname{section}{Section}{Sections}
\crefname{equation}{Eq.}{Eqs.}
\Crefname{equation}{Equation}{Equations}
\crefname{table}{Table}{Tables}
\Crefname{table}{Table}{Tables}

\newcommand*{\new}{\textcolor{Black}}

\usepackage{ifthen}
\newboolean{showcomments}
\setboolean{showcomments}{false}
\ifthenelse{\boolean{showcomments}}
{ \newcommand{\mynote}[3]{
		\fbox{\bfseries\sffamily\scriptsize#1}
		{\small$\blacktriangleright$\textsf{\emph{\color{#3}{#2}}}$\blacktriangleleft$}}
	\newcommand{\zzz}[1]{{\setlength{\fboxsep}{2pt}\fcolorbox{black}{yellow}{\textsf{\emph{#1}}}}\xspace}}
{ \newcommand{\mynote}[3]{}
	\newcommand{\zzz}[1]{}}

\usepackage{amsmath,amsfonts,bm}

\def\eqref#1{equation~\ref{#1}}

\def\1{\bm{1}}

\definecolor{baselinecolor}{RGB}{26, 128, 187}
\newcommand{\baselinecolor}{baselinecolor}
\definecolor{speccolor}{RGB}{187, 26, 129}
\newcommand{\speccolor}{speccolor}

\newcommand{\add}[1]{\scriptsize{\textcolor{Green}{#1}}}
\newcommand{\sub}[1]{\scriptsize{\textcolor{Red}{#1}}}

\definecolor{bleuhtml}{HTML}{1a80bb}
\definecolor{yellowhtml}{HTML}{f2c45f}
\definecolor{purplehtml}{HTML}{bb1a81}

\def\cls{{[\texttt{CLS}]\xspace\xspace}}

\DeclareMathAlphabet{\mathsfit}{\encodingdefault}{\sfdefault}{m}{sl}
\SetMathAlphabet{\mathsfit}{bold}{\encodingdefault}{\sfdefault}{bx}{n}

\title{Revisiting \cls and Patch Token Interaction \\ in  Vision Transformers}

\author{Alexis Marouani\textsuperscript{1,2}\thanks{Correspondence to \texttt{amarouani@meta.com}} \And Oriane Siméoni\textsuperscript{1} \And Hervé Jégou\textsuperscript{1} \And Piotr Bojanowski\textsuperscript{1} \And Huy V. Vo\textsuperscript{1}
\AND \\
\textsuperscript{1} FAIR, Meta \\
\textsuperscript{2} LIGM, Ecole des Ponts, IPParis, UGE, CNRS, 77455 Marne-la-Vallée, France \\
}

\iclrfinalcopy
\begin{document}

\maketitle

\begin{abstract}
Vision Transformers have emerged as powerful, scalable and versatile representation learners. To capture both global and local features, a learnable \cls class token is typically prepended to the input sequence of patch tokens. Despite their distinct nature, both token types are processed identically throughout the model.
In this work, we investigate the friction between global and local feature learning under different pre-training strategies by analyzing the interactions between class and patch tokens.
Our analysis reveals that standard normalization layers introduce an implicit differentiation between these token types. Building on this insight, we propose specialized processing paths that selectively disentangle the computational flow of class and patch tokens, particularly within normalization layers and early query-key-value projections.
This targeted specialization leads to significantly improved patch representation quality for dense prediction tasks. Our experiments demonstrate segmentation performance gains of over 2 mIoU points on standard benchmarks, while maintaining strong classification accuracy. The proposed modifications introduce only an 8\% increase in parameters, with no additional computational overhead.
Through comprehensive ablations, we provide insights into which architectural components benefit most from specialization and how our approach generalizes across model scales and learning frameworks.

\end{abstract}
\section{Introduction}
\label{sec:intro}

\vspace{-5pt}
\begin{figure}[b!]
    \centering
\begin{subfigure}[b]{0.195\textwidth}
\includegraphics[width=\textwidth]{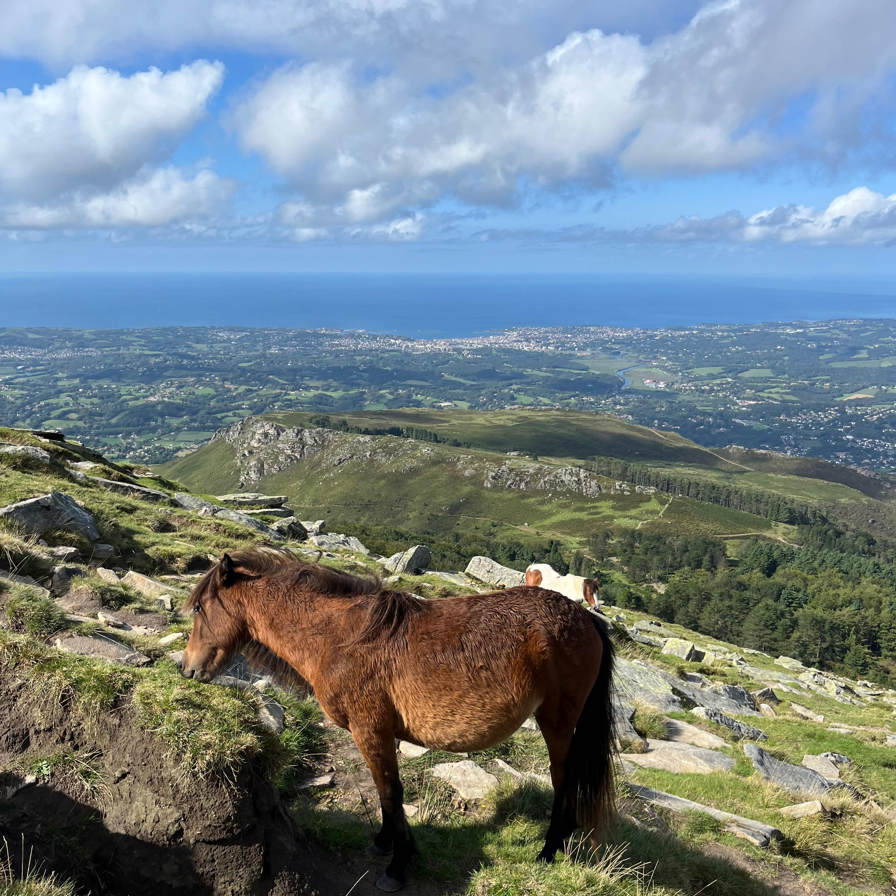}
\caption*{Original image \\ \textcolor{white}{l}}
\end{subfigure}
\hfill
\begin{subfigure}[b]{0.195\textwidth}
\includegraphics[width=\textwidth]{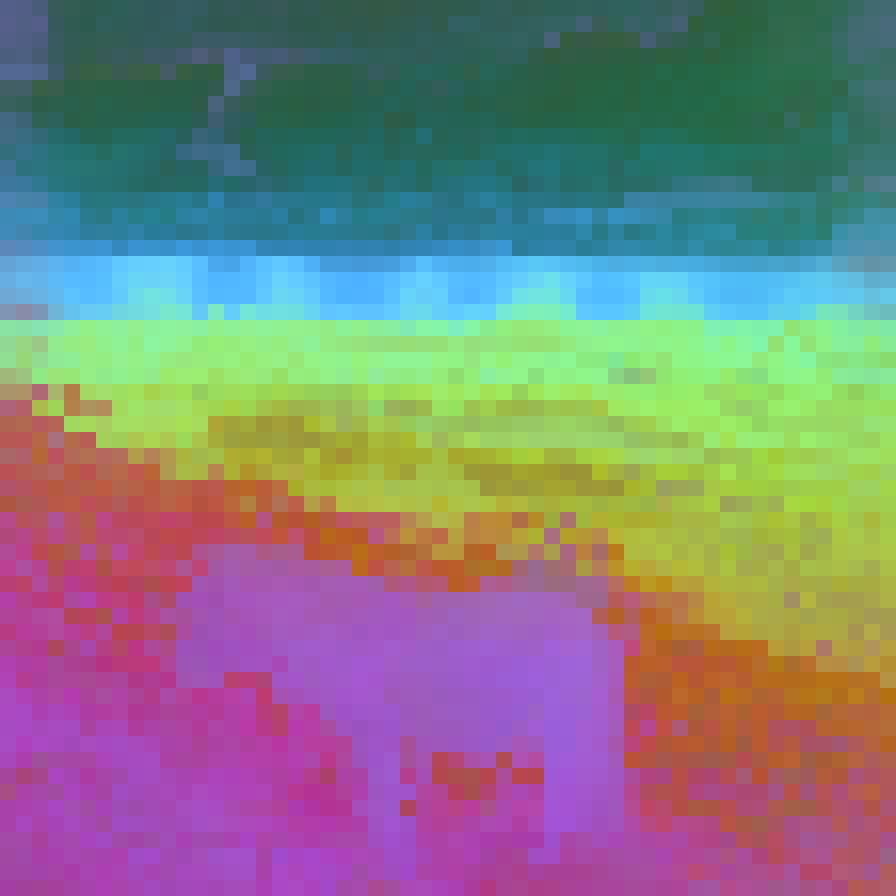}
\caption{DINOv2 w/ regs \\ \citep{darcet2023vision}}
\end{subfigure}
\hfill
\begin{subfigure}[b]{0.195\textwidth}
\includegraphics[width=\textwidth]{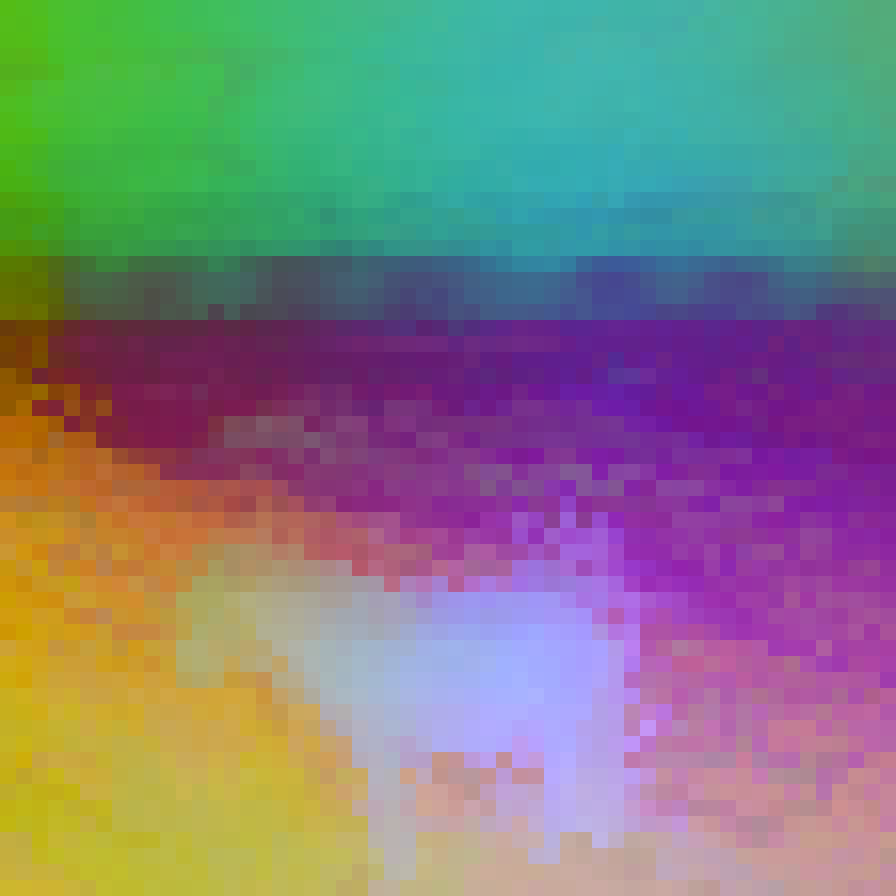}
\caption*{(a) + ours \\ \textcolor{white}{l}}
\end{subfigure}
\hfill
\begin{subfigure}[b]{0.195\textwidth}
\includegraphics[width=\textwidth]{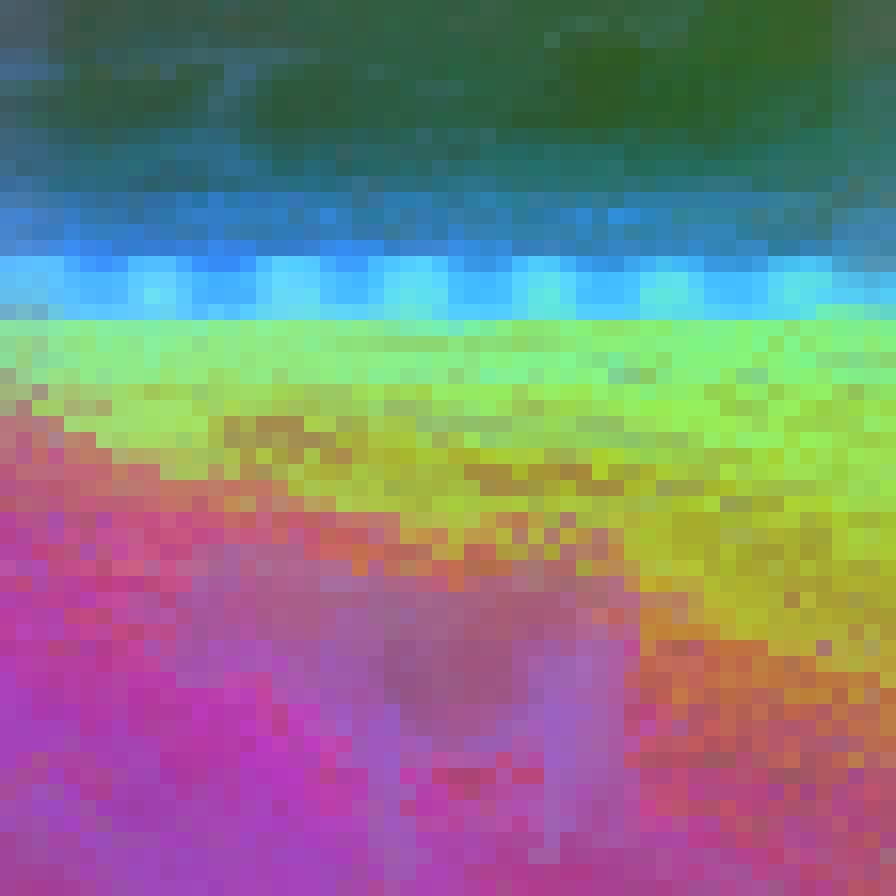}
\caption{DINOv2 w/ attn. \\  bias \citep{an2025systematic}}
\end{subfigure}
\hfill
\begin{subfigure}[b]{0.195\textwidth}
\includegraphics[width=\textwidth]{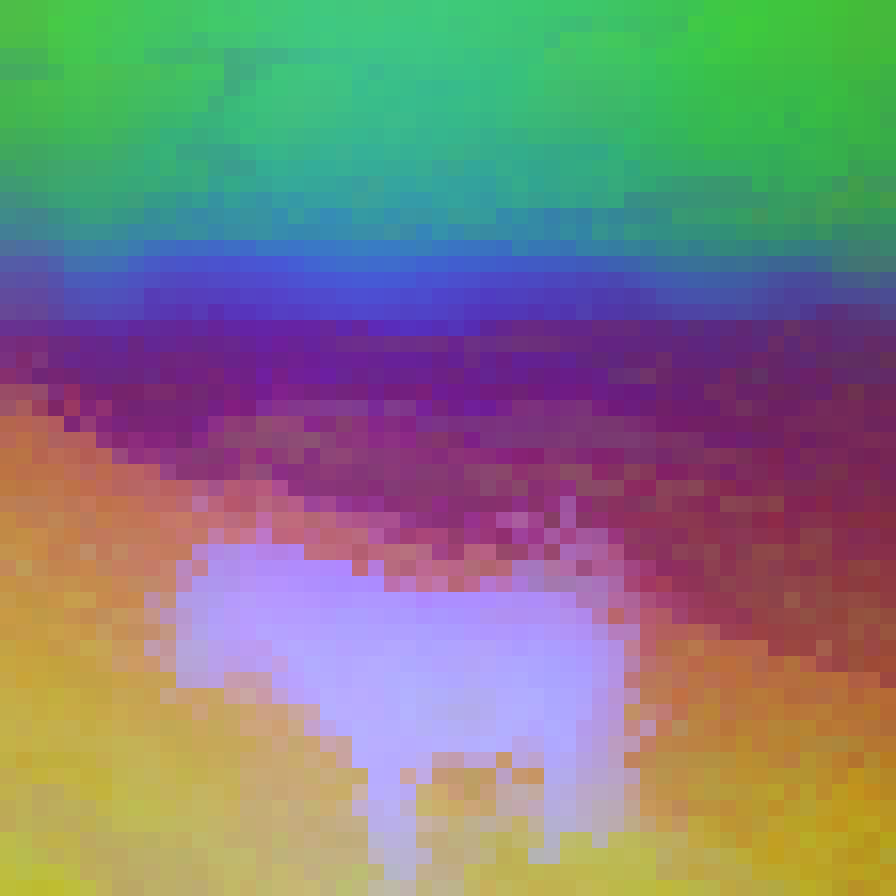}
\caption*{(b) + ours \\ \textcolor{white}{l}}
\end{subfigure}
\hfill

\vspace{-15pt}
\caption{\textbf{Visualization of the impact} of our proposed layer specialization for \cls and patch tokens on the patch features obtained with DINOv2 when using two strategies to mitigate artifacts, namely registers (`regs') \citep{darcet2023vision} and attention bias (`attn. bias') \citep{an2025systematic}. We display the first PCA components of model outputs in RGB.}
\label{fig:intro}
\end{figure}

In recent years, significant progress has been made in developing vision foundation models capable of generating 
rich and highly generalizable visual representations for images. Notably, latest state-of-the-art results have been achieved 
using Vision Transformer (ViT) models~\citep{dosovitskiy2020image} trained under various paradigms, including 
fully-supervised \citep{touvron2022deit}, weakly supervised \citep{radford2021learning, bolya2025perception}, and self-supervised learning 
\citep{zhou2021ibot, oquab2023dinov2, simeoni2025dinov3}. These models capture a wide spectrum of visual semantics, enabling robust 
performance across a diverse range of downstream tasks and data domains.

The ViT architecture~\citep{dosovitskiy2020image} processes images by dividing them into fixed-size patches, 
which are then embedded and fed to a sequence of transformer blocks. Typically, a trainable class token \cls is prepended to the 
sequence of patch embeddings and is designed to aggregate information from all patches. Patches and \cls tokens are  trained with 
different objectives, if any. For instance, most pre-training methods apply a loss function solely on the \cls token 
~\citep{chen2020simple,grill2020bootstrap, caron2021emerging,radford2021learning, touvron2022deit}. Some employs an objective on patch tokens 
only \citep{he2022masked}, while others train the \cls and patch tokens with separate losses \citep{zhou2021ibot, oquab2023dinov2, simeoni2025dinov3}.
Regardless of the specific training paradigm, recent works~\citep{darcet2023vision, an2025systematic, simeoni2025dinov3} show that there is a persistent 
imbalance between the \cls and patch tokens.
Proposed solutions to this issue include introducing additional storage tokens into the input sequence~\citep{darcet2023vision}, modifying 
the attention mechanism~\citep{an2025systematic}, or incorporating additional loss terms to explicitly constrain patch locality~\citep{simeoni2025dinov3}.
In contrast, we hypothesize that the imbalance arises because models process the \cls and patch tokens through identical computational 
pipelines, despite their fundamentally different roles and nature, and propose disentangling their treatment to overcome the imbalance.

In this work, we analyze the model statistics in order to better understand the internal mechanisms that govern the interaction between the \cls and patch tokens. 
Our analysis reveals a surprising finding: normalization layers are already implicitly learning to distinguish between the \cls and patch 
tokens before the attention mechanism. Building on this insight, we introduce a simple yet effective architectural modification that explicitly 
separates the processing of the \cls and patch tokens, as illustrated in \cref{fig:architecture}. 
With just a minimal set of specialized layers, our approach leads to noticeably richer dense features (see \cref{fig:intro}) and delivers substantial 
gains on dense prediction tasks. For instance, we improve the average mIoU scores on segmentation benchmarks by as much as 2.2 points with a ViT-L. 
This work sheds light on hidden dynamics within transformer models and also demonstrates how targeted architectural changes can translate into significant 
real-world performance improvements. We make the following contributions:

\begin{itemize}
    \item We analyze the interactions between \cls{} and patch tokens within Vision Transformers, and show that models implicitly attempt to distinguish them through normalization layers.
    \item We propose an architectural modification that specializes their computations to reduce the friction between them, while keeping the number of operations constant. We study different specialization strategies for transformer block components.
    \item We demonstrate the generalizability of our 
approach across model scales and learning frameworks, showing significant improvements in dense prediction 
tasks without compromising classification performance. 
\end{itemize}
\section{Related Work}
\label{sec:related}

\paragraph{Vision Transformers}
Inspired by \citet{vaswani2017attention} and first introduced by \citet{dosovitskiy2020image}, Vision Transformer has become 
an architecture of choice when building vision models.
A typical ViT model consists of a patch embedder and a stack of transformer blocks. Given an image, the patch embedder 
divides it into equally-sized square patches and transforms them into patch tokens that represent local information in the image.
Optionally, a learnable \cls token is added to the set of patch tokens in order to capture global information. All tokens are then 
passed through the transformer blocks which process them with various transformations, most notably the multi-head self-attention operator 
\citep{vaswani2017attention} that allows tokens to attend to each other.
Built on the original architecture, subsequent works have introduced additional components to improve various aspects of ViTs such as 
data efficiency \citep{touvron2020training,yuan2021tokens}, computational cost \citep{Liu2021swin,bolya2023token} and normalization \citep{touvron2021going}.
ViT architecture has enabled state-of-the-art performance in various tasks \citep{carion2020detr,Strudel_2021_ICCV}, simplified multi-modal 
learning \citep{radford2021learning,fini2024multimodal}, and led to excellent local and global representation in foundation models 
\citep{oquab2014learning,tschannen2025siglip,simeoni2025dinov3}.  
In most ViTs, \cls and patch tokens are functionally interchangeable in transformer blocks -- they are processed in identical manner 
using the same operators -- despite their distinctive nature. We show in our analysis that the identical treatment of these tokens is suboptimal 
and that disentangling them leads to better local features for dense tasks.

\paragraph{Improving dense feature learning}
Visual representation learning approaches have mostly focus on optimizing the global representation by primarily training 
the \cls token to summarize the image content either in supervised \citep{touvron2022deit}, weakly supervised \citep{radford2021learning,tschannen2025siglip} 
or self-supervised settings \citep{caron2021emerging,Liu2021swin}. As a by-product, they also produce local representation that perform well on tasks 
that require fine-grained  features such as object detection, semantic segmentation or depth estimation.
Most notably, the self-supervised method DINO \citep{caron2021emerging} produces excellent patch features that supercharge research on unsupervised 
object detection and segmentation.
iBoT \citep{zhou2021ibot} augments DINO with masked image modeling \citep{he2022masked} to optimize both global and local representation.
DINOv2 \citep{oquab2023dinov2} introduces new technical components such as Sinkhorn-Knopp centering and untying heads to 
successfully scale DINO to large datasets and model sizes, achieving excellent performance on dense tasks.
Learning meaningful dense features with Vision Transformers is not without challenges.
\citet{darcet2023vision} discusses 
the noisy attention maps produced by models trained at scale during longer training periods.
This issue, which degrades dense prediction performance, is caused by some patch tokens losing their local context after being repurposed by the model to store global information. 
They propose an architectural solution with registers to mitigate these issues.
Other successful attempts to enhance the quality of local features include regularizing similarity to neighbor patches post-training 
\citep{pariza2024near} or recovering patch similarity with Gram anchoring mechanism \citep{simeoni2025dinov3}. Similar to these works, 
we improve dense features quality during training by specializing \cls and patch tokens treatment within the Transformer blocks of 
ViTs and thus reducing the friction between them.

\section{Friction Between \cls and Patches}
\label{subsec:cls-patch-modalities}

\begin{figure}[t]
    \centering
    \begin{subfigure}[b]{0.24\textwidth}
        \includegraphics[height=2.25cm]{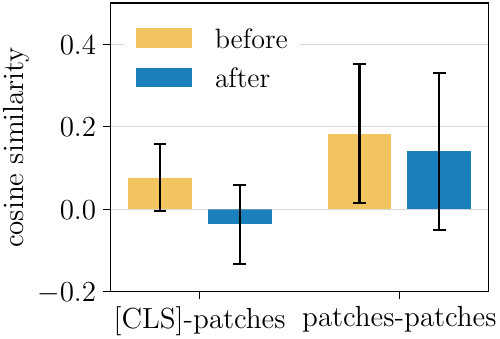}
        \caption{LayerNorm pre-attn.}
    \end{subfigure}
    \begin{subfigure}[b]{0.24\textwidth}
        \includegraphics[height=2.25cm]{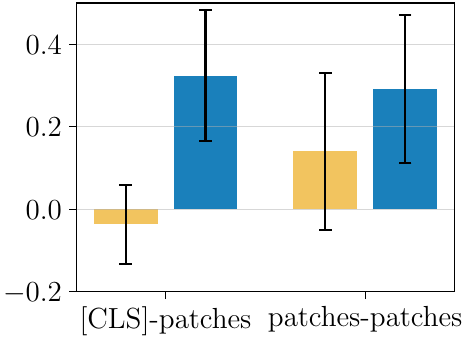}
        \caption{Self-attention}
    \end{subfigure}
    \begin{subfigure}[b]{0.24\textwidth}
        \includegraphics[height=2.25cm]{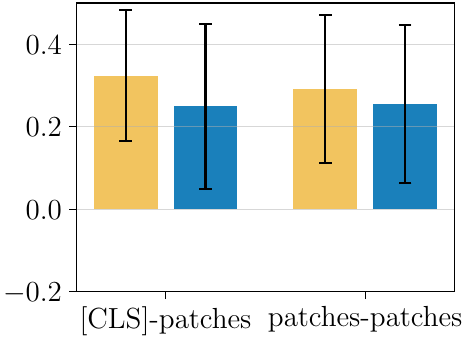}
        \caption{LayerScale post-attn.}
    \end{subfigure}
    \begin{subfigure}[b]{0.24\textwidth}
        \includegraphics[height=2.25cm]{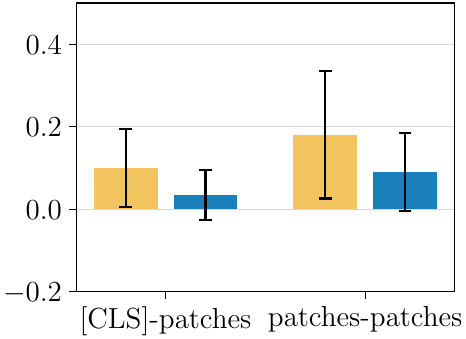}
        \caption{LayerNorm pre-MLP}
    \end{subfigure}
    \setlength{\fboxsep}{0pt} 
    \caption{\textbf{\cls-patches separation effect within transformer blocks}
    in vanilla DINOv2 ViT-L model. We show mean and standard deviation of cosine similarity between \cls and all patches, and all-to-all patches, \colorbox{yellowhtml}{\strut before} and \colorbox{bleuhtml}{\strut\textcolor{white}{after}} each transformer layers. `attn.' stands for attention.}
    \label{fig:layers-effect}
    \vspace{-10pt}
\end{figure}

Vision Transformers are typically trained with a trainable \cls token which encodes global information about the image prepended to the sequence of patch tokens.  
Despite the distinct nature of the \cls and patch tokens, current models treat them equivalently, 
applying the exact same operations to both. However, \citep{darcet2023vision} has highlighted potential communication issues 
between these two types of tokens, leading to a severe loss of locality of patch tokens and the appearance of undesirable outliers 
in the attention maps. While registers help to mitigate the appearance of artifacts, we argue that more could be done. 
Our observations indicate that a degree of friction persists between \cls{} and patch tokens, as discussed below.

\paragraph{ViTs differentiate \cls and patch tokens for the attention} We analyze the interplay between \cls and patch tokens by computing their similarity at different points within the model, before and after principal layers in each transformer block. 
In \cref{fig:layers-effect}, we visualize the mean and standard deviation of these similarities. 
Our results are averaged over patches of $1000$ images and across all model blocks. 
Additionally, we present the same statistics between patches.
While certain operations—such as the LayerScale applied post-attention—have little effect 
on the similarity between \cls{} and patch tokens, the self-attention layer markedly increases their similarity. 
This increase is expected, as self-attention realigns the different token types. However, our analysis uncovers a surprising phenomenon: 
the representations of \cls{} and patch tokens naturally diverge at specific stages of the computational pipeline, particularly just 
before attention operations. Indeed, the LayerNorm applied before attention drastically reduces the similarity between \cls{} and patch tokens, bringing it close to zero.
This implicit differentiation indicates that the model attempts to adapt to the distinct functional roles of these token types before the attention mechanism, 
despite their shared parameterization.
We plot the statistics of more layers in \cref{ap:additional-layers-effect}.

\begin{figure}[t]
    \centering
    \begin{subfigure}[b]{0.24\textwidth}
        \includegraphics[height=2.25cm]{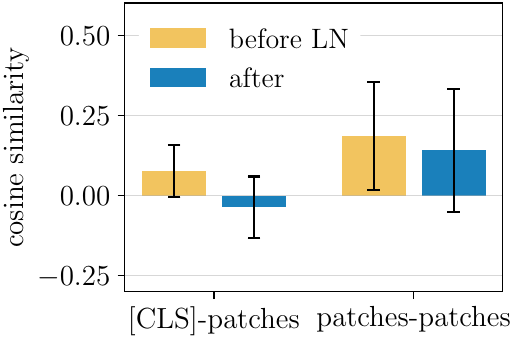}
        \caption{DINOv2}
        \label{subfig:layernorm_effect-dinv2}
    \end{subfigure}
    \begin{subfigure}[b]{0.24\textwidth}
        \includegraphics[height=2.25cm]{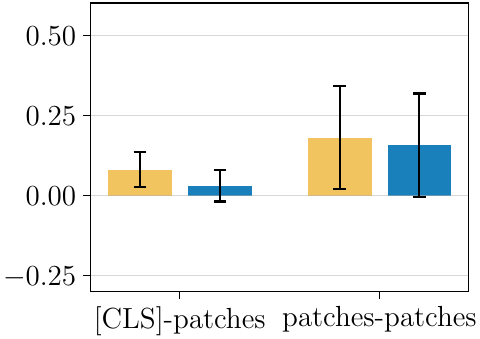}
        \caption{DINOv2 w/ regs}
        \label{subfig:layernorm_effect-dinv2-reg}
    \end{subfigure}
    \begin{subfigure}[b]{0.24\textwidth}
        \includegraphics[height=2.25cm]{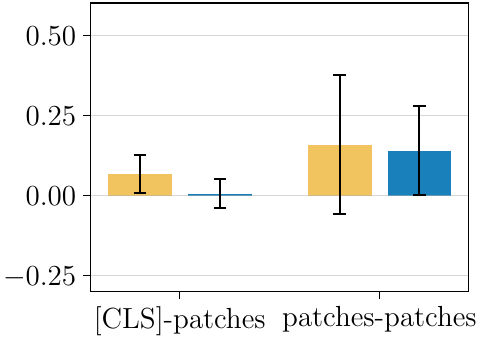}
        \caption{DINOv2 w/ attn-bias}
        \label{subfig:layernorm_effect-dinv2-attbias}
    \end{subfigure}
    \begin{subfigure}[b]{0.24\textwidth}
        \includegraphics[height=2.25cm]{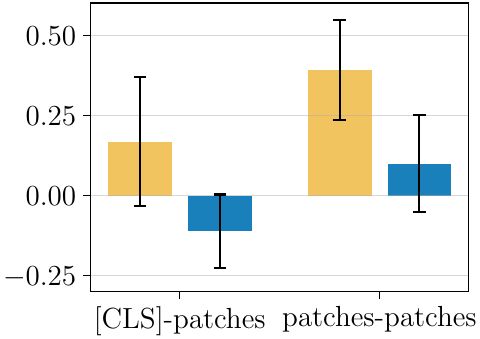}
        \caption{DEIT-III}
        \label{subfig:layernorm_effect-deit}
    \end{subfigure}
    \setlength{\fboxsep}{1pt} 
    \caption{\textbf{Impact of LayerNorm before attention layer}
    for different pre-trained models. We show mean and standard deviation of cosine similarity between \cls and all patches, and between all patches. Statistics visualized \colorbox{yellowhtml}{before} and \colorbox{bleuhtml}{\textcolor{white}{after}} LayerNorm (LN).}
    \label{fig:layernorm-effect}
    \vspace{-15pt}
\end{figure}

\paragraph{Role of the pre-attention LayerNorm}
In \cref{fig:layernorm-effect}, we focus on the impact of the pre-attention LayerNorm to the similarity between the \cls and patch tokens in different pre-trained models, 
including DINOv2~\citep{oquab2023dinov2} and its variants with registers~\citep{darcet2023vision}, noted `regs', and attention bias~\citep{an2025systematic}, noted `attn. bias', and supervised DEIT-III~\citep{touvron2022deit}.
It can be observed that in all cases, prior to the attention mechanism, the LayerNorm disentangles the \cls and patch tokens, enabling them to serve distinct functions 
within the attention process.
This phenomenon appears in all pre-trained models with different extent.
For instance, the LayerNorm strongly enforces negative correlation between \cls and patches in DINOv2 and DEIT-III while keeping the correlation close 
to zeros in the variants of DINOv2.
In contrast, the similarity among patch tokens remains positive and largely stable, with only a slight decrease observed---a phenomenon we interpret as a regularization effect. This effect likely prevents rank collapse and promotes a more uniform distribution of tokens on the unit sphere, consistent with 
observations reported in \citet{wu2024role}.

\begin{figure}[h]
\centering
\begin{subfigure}[b]{0.3\textwidth}
\includegraphics[width=\textwidth, height=3.5cm]{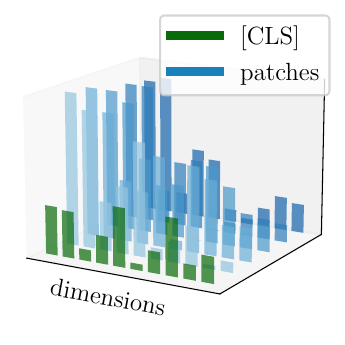}
\caption{Block 7}
\end{subfigure}
\hfill
\begin{subfigure}[b]{0.3\textwidth}
\includegraphics[width=\textwidth, height=3.5cm]{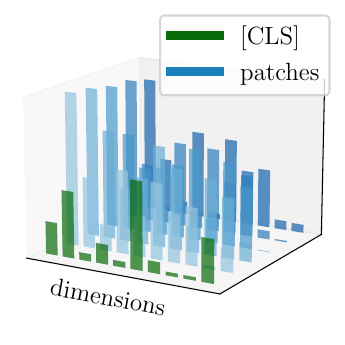}
\caption{Block 15}
\end{subfigure}
\hfill
\begin{subfigure}[b]{0.3\textwidth}
\includegraphics[width=\textwidth, height=3.5cm]{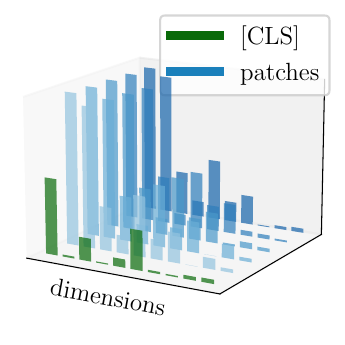}
\caption{Block 23 (Last)}
\label{fig:3d_blk23}
\end{subfigure}
\hfill
\begin{subfigure}[b]{\textwidth}
\end{subfigure}
\caption{\textbf{Dimensions with biggest magnitudes} early (a), in the middle (b), at the end (c) of the model for \cls \new{and $5$ patches with the highest magnitudes in the selected dimensions}. Tokens taken at the output of blocks. The considered model is a DINOv2 ViT-L with attention bias.}
\label{fig:3d}
    \vspace{-15pt}
\end{figure}

\paragraph{Dimension separation} \new{To understand how a separation effect can appear in a LayerNorm layer, one has to recall that it performs a point-wise normalization and a dimension-wise affine transformation. Therefore, a separation effect occurs when inputs have very different magnitudes in each dimension.
In \cref{fig:3d}, we plot the dimensions with biggest absolute magnitudes---averaged over patches and \texttt{[CLS]}---at the output of different blocks. 
We observe that some specific dimensions are leveraged only by a certain token type. For example, in \cref{fig:3d_blk23}, the $2$nd dimension presents large magnitudes for patches and almost none for \cls. Moreover, the deeper we are in the model, the fewer token types share dimensions.}
This enables normalization layers to perform distinctive operations. More than just regularizing, they specialize and separate the tokens.

All the observations above indicate that treating \cls and patch tokens identically compels the model to allocate resources towards implicitly 
separating them,  
which could be used to learn more meaningful features. 
We argue that disentangling their treatments would facilitate the model in learning 
better representation, as discussed in next section.

\section{\cls - Patches Specialization: Analysis }

In this section, we first define our proposed layer specialization in \cref{sec:proposal} and set the experimental setting in \cref{sec:exp-setting}.  
In \cref{subsec:norm}, we discuss the benefits of splitting normalizations for \cls and patch tokens.
We also investigate which part of the model needs specialization in \cref{subsec:depth}, and more specifically which layers in \cref{subsec:layers}. 

\subsection{Our Proposal: Layer Specialization}
\label{sec:proposal}

\begin{figure}[t]
\centering
\includegraphics[width=\textwidth]{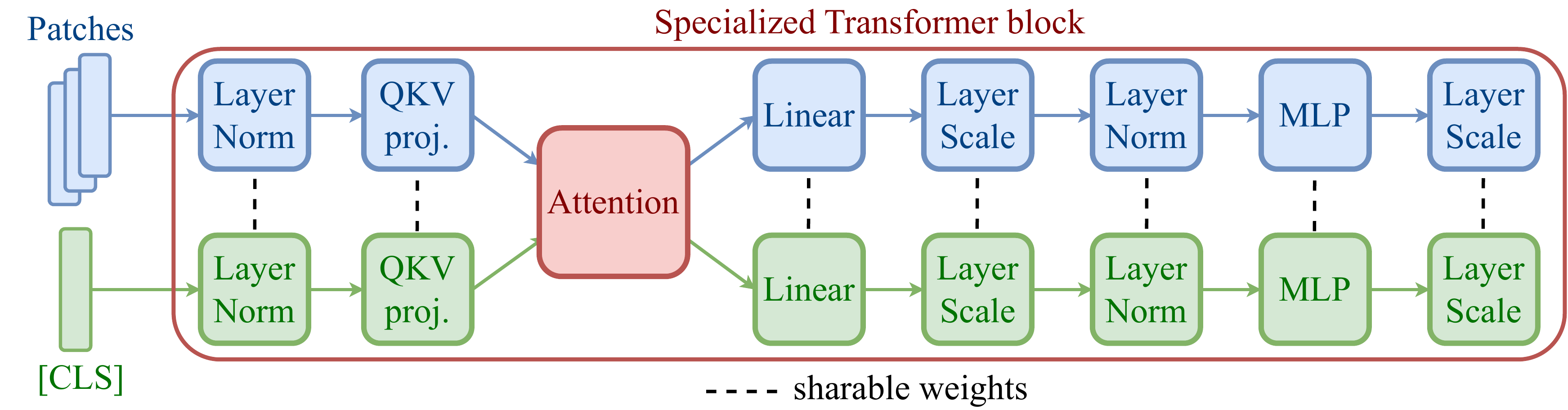}
\vspace{-15pt}
\caption{\textbf{Architecture specialization.}  We investigate how \cls and patch tokens can be processed through specialized layers, while preserving their interactions within the attention mechanism.}
\label{fig:architecture}
\vspace{-10pt}
\end{figure}

Based on observations made in the previous section, we explore disentangling the computation of global and local representations in ViTs.
Taking inspiration from the success of double-stream architectures to handle different modalities \citep{esser2024scaling}, we explore a similar approach for the \cls and patch tokens.
More specifically, inside a classic transformer block, \cls and patch tokens
go through several layers: projections, some normalizations and a MLP.
We propose to decouple the \cls and patch tokens by processing them with different weights for certain layers.
Indeed, instead of using a single layer to process both token types, we introduce two distinct layers---each with its own set of weights---specialized for either \cls or patch tokens. This allows each layer to better capture the unique characteristics of its respective token type.
However, the tokens continue to interact through the attention mechanism as usual, ensuring information flow between \cls and patch tokens is preserved. 
An illustration of this specialized architecture is provided in \cref{fig:architecture}.
While this approach introduces some additional memory overhead, our experiments show that the increase in model size remains small---approximately 8\%---when layer specialization is applied selectively to achieve optimal performance.
More importantly, layer specialization does not increase inference FLOPs, as the model continues to perform the same computational operations during inference. This ensures that the efficiency of the model is maintained, even as we enhance its representational capacity through targeted specialization.

\subsection{Experimental Setting: Training and Evaluation}
\label{sec:exp-setting}

\paragraph{Training}
We investigate layer specialization with different pre-training paradigms including
the popular self-supervised strategy DINOv2 \citep{oquab2023dinov2} and the fully-supervised DeiT-III \citep{touvron2022deit}. We also investigate different model sizes (ViT-B, L, H). 
Unless specified otherwise, we produce results with a ViT-L model trained following DINOv2 recipe. 
Following \citet{an2025systematic}, we integrate the attention bias strategy, which mitigates high-norm anomalies \citep{darcet2023vision} without introducing additional tokens, in all models and attention operations. 
More discussion can be found in \cref{ap:anomalies}.
For DINOv2, we train our models on ImageNet-$22$K~\citep{ridnik2021imagenet} dataset for $600$k training steps.
For DeiT-III, we train our models on ImageNet-$1$K~\citep{deng2009imagenet} for respectively $400$ epochs on ViT-B and $800$ epochs on ViT-L.
For both training paradigms, we pre-train using the first pre-training phase, and drop the high-resolution fine-tuning step. 
We report more details in \cref{ap:eval-details}.

\paragraph{Evaluation}
Following \citet{oquab2014learning}, we assess model representations via linear probing on global, with ImageNet-1k \citep{deng2009imagenet}, and dense prediction tasks. For semantic segmentation, we use ADE20K~\citep{zhou2017scene}, Cityscapes~\citep{cordts2016cityscapes} and  PASCAL VOC~\citep{everingham2010pascal}, reporting mIoU. For depth estimation, we use KITTI~\citep{geiger2013vision}, NYU Depth v2~\citep{Silberman:ECCV12} and SUN RGB-D~\citep{song2015sun}, reporting RMSE.
\new{For detection, we use COCO \citep{lin2014microsoft}, reporting AP.}
Some tables show average segmentation and depth scores across corresponding benchmarks.
More details in
\cref{ap:eval-details}.

\begin{figure}[t]
\centering
\begin{subfigure}[b]{0.3\textwidth}
    \centering
    \includegraphics[height=2.6cm]{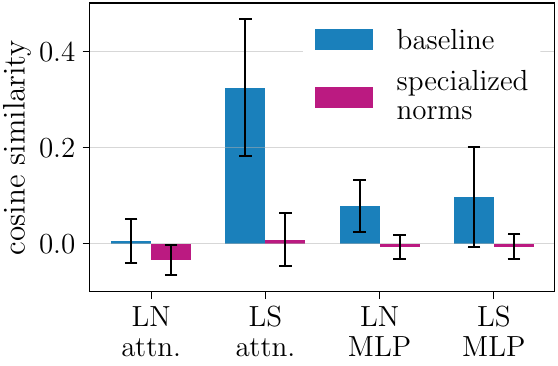}
    \vspace{-5pt}
    \caption{\cls-patches cosine sim.}
    \label{subfig:spec_impact_cls}
\end{subfigure}
\hfill
\begin{subfigure}[b]{0.3\textwidth}
    \centering
    \includegraphics[height=2.6cm]{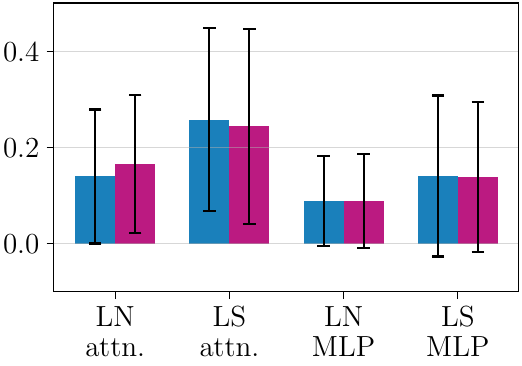}
    \vspace{-5pt}
    \caption{Patches-patches cosine sim.}
    \label{subfig:spec_impact_rdm}
\end{subfigure}
\hfill
\begin{subfigure}[b]{0.35\textwidth}
    \resizebox{\linewidth}{!}{
    \begin{tabular}{l ccc}
        \toprule
         Spec. & Linear &  Avg  & Avg\\
        & Acc.  &  Seg.  & Depth $\downarrow$ \\
        \midrule
         -- & \bf 85.4 & 64.5 & 1.232 \\
         norms  & 85.1 & \bf 65.6 & \bf 1.178 \\
        \bottomrule
    \end{tabular}
    }
    \caption{Norm specialization}
    \label{subtab:norm-spec-res}
\end{subfigure}
\vspace{-5pt}
\setlength{\fboxsep}{1pt} 
\caption{\textbf{Normalization specialization.} Mean and 
standard deviation of the cosine similarity computed between (a) \cls and all patches and (b) all to all patches. 
We 
compare post-normalization statistics between the \colorbox{bleuhtml}{\textcolor{white}{baseline}} architecture 
(DINOv2 ViT-L with attn. bias) and when specializing the normalization layers 
for \cls and patch token (\colorbox{purplehtml}{\textcolor{white}{specialized norms}}). 
(c) Quantitative results with specialized normalization layers. `LN' stands for LayerNorm and `LS' for 
LayerScale.}
\label{fig:norm_role}
\vspace{-10pt}
\end{figure}

\subsection{Specializing Normalization Layers}\label{subsec:norm}

As discussed in \cref{subsec:cls-patch-modalities}, ViTs attempt to separate the \cls and patch tokens with the LayerNorm applied prior to the attention operation. Building on this observation, our initial experiment focuses on specializing the normalization layers (LayerNorms and Layer Scales) within the model, with the aim of further supporting the model’s inherent tendency to separate these feature types.

We specialize the normalization layers in all blocks of the model. 
This lightweight modification introduces only 0.05\% additional parameters, yet significantly alters the feature distributions. 
In \cref{subfig:spec_impact_cls}, we report the mean and standard deviation of the cosine similarity between the \cls and patch tokens, computed after each normalization layer. We compare our variant with specialized normalization weights to the baseline.
Conversely, \cref{subfig:spec_impact_rdm} shows the corresponding statistics when using all patches instead of the \cls. We observe that specializing the normalization layers further amplifies the disentanglement of the \cls and patch tokens, resulting in a more distinct separation of their embeddings after each normalization step. 

The impact of these specialized normalizations is quantified in \cref{subtab:norm-spec-res}. The specialization leads to significant improvements on dense prediction tasks, yielding an average increase of $+1.1$ mIoU points on segmentation benchmarks and an improvement of $-0.054$m on depth estimation. These results highlight that a better specialization of the token types benefits the patches representations. On the other side, global results are slightly degrading. We however show in the next section that this loss can be mitigated. Unless otherwise specified, in the remainder of the paper, we apply specialized normalization layers to all transformer blocks.

\subsection{Block-Level Targeted Specialization}
\label{subsec:depth}

\begin{wrapfigure}[17]{r}{6.5cm}
    \centering
    \includegraphics[width=0.8\linewidth]{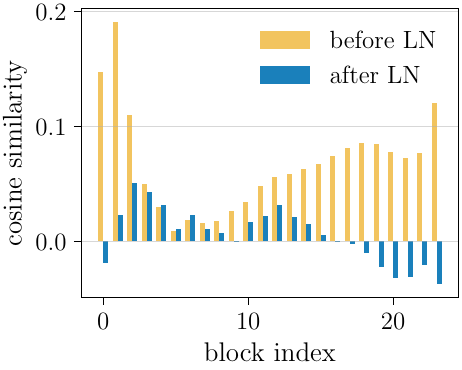}
    \caption{Mean cosine similarity between \cls and all patches \colorbox{yellowhtml}{before} and \colorbox{bleuhtml}{\textcolor{white}{after}} the first LayerNorm (LN) of each block.}
    \label{fig:normlayers}
\end{wrapfigure}

While normalization layers in ViTs show in overall a \cls-patch separation effect, we have observed that the extent of their impact is not uniform across all blocks. 
It can be seen from \cref{fig:normlayers}, which depicts \cls-patches cosine similarity before and after the first LayerNorm in each block, that the separation effect of the normalization varies depending on its position within the model. Indeed,  blocks at the beginning and near the end see the most impact.
We hypothesize that the importance of separation in the early blocks stems from their proximity to the different inputs.
Although the \cls token is trained to summarize information from the patches, it is initialized as a learned parameter and thus has an input distribution very distinct from that of the patch tokens.
Later in the model, separation becomes important again as tokens are closer to the final representations and the training objectives. The observations above suggest that we can benefit from more targeted specialization within the model. We study next which blocks should be specialized to optimize the model's performance. 

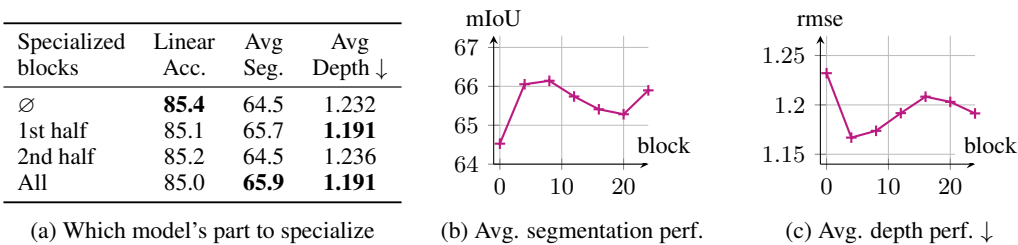
\begin{figure}[t]
\centering
    \begin{subfigure}[b]{0.39\textwidth}
    \centering
    \resizebox{\textwidth}{!}{
    \begin{tabular}{l ccc}
        \toprule
        Specialized & Linear &  Avg  & Avg\\
        blocks & Acc.  &  Seg.  & Depth $\downarrow$ \\
        \midrule
        $\varnothing$ & \textbf{85.4} & 64.5 & 1.232 \\
         1st half   & 85.1 & 65.7 &\bf 1.191 \\
         2nd half  & 85.2 & 64.5 & 1.236 \\
         All
         & 85.0 & \textbf{65.9} & \bf 1.191 \\
        \bottomrule
    \end{tabular}
    }
    \caption{Which model's part to specialize}
    \label{tab:whichpart}
    \end{subfigure}
    \hfill
    \begin{subfigure}[b]{0.29\textwidth}
    \centering
    \begin{tikzpicture}
\begin{axis}[
    scale=0.3,
    every node/.style={scale=1},
    scaled x ticks=false,
    xlabel={block},
    ylabel={mIoU},
    ymajorgrids,
    xmajorgrids,
    tick label style={font=\small},
    label style={font=\small},
    legend cell align={left},
    legend pos=north east,
    ylabel near ticks,
    xmin=-1, 
    ymin=64,
    ymax=67.3,
    legend style={nodes={scale=0.6, transform shape}, draw=none},
    axis lines=left,
    x label style={at={(axis description cs:1.1,0.18)},anchor=north, fill=white,font=\small,yshift=5pt},
    y label style={at={(axis description cs:0.0,0.9)},anchor=south, rotate=-90, font=\small,yshift=5pt},
]
\addplot[mark=+, draw=\speccolor, thick] 
    table [x=depth, y=seg_performances_sorted, col sep=comma] {figures/numbers/perfs_vs_split_depth/perfs_vs_depth.csv}; 

\end{axis}
\end{tikzpicture}
\caption{Avg. segmentation perf.}
\label{fig:depth-seg}
    \end{subfigure}
    \hfill
    \begin{subfigure}[b]{0.29\textwidth}
    \centering
       \begin{tikzpicture}
        \begin{axis}[
            scale=0.3,
            every node/.style={scale=1},
            scaled x ticks=false,
            xlabel={block},
            ylabel={rmse},
            ymajorgrids,
            xmajorgrids,
            tick label style={font=\small},
            label style={font=\small},
            legend cell align={left},
            legend pos=north east,
            ylabel near ticks,
            xmin=-1, 
            ymin=1.14,
            ymax=1.27,
            legend style={nodes={scale=0.6, transform shape}, draw=none},
            axis lines=left,
            x label style={at={(axis description cs:1.1,0.18)},anchor=north, fill=white,font=\small,yshift=5pt},
            y label style={at={(axis description cs:0.0,0.9)},anchor=south, rotate=-90, font=\small,yshift=5pt},
        ]
        \addplot[mark=+, draw=\speccolor, thick] 
            table [x=depth, y=depth_performances_sorted, col sep=comma] {figures/numbers/perfs_vs_split_depth/perfs_vs_depth.csv}; 

        \end{axis}
    \end{tikzpicture}
    \caption{Avg. depth perf. $\downarrow$}
\label{fig:depth-depth}
    \end{subfigure}
    \vspace{-5pt}
\caption{\textbf{Block specialization.} (a) Performance metrics when specializing specific parts of the model (b) Average segmentation scores and (c) average depth rmse ($\downarrow$) vs number of specialized blocks at the beginning of the model. Normalization layers are specialized in all blocks. The base model is a DINOv2 ViT-L with attention bias.}
\label{fig:depth_influence}
\vspace{-10pt}
\end{figure}

We first quantitatively compare the impact of specializing different sections of the model, as shown in \cref{tab:whichpart}.
To this end, we train DINOv2 while specializing either the first half, the second or all of the transformer blocks, on top of specializing all normalization layers.
Within a block, all layers are specialized.
Our findings indicate that the best performance is achieved when specializing the early layers, which are closest to the input. Specifically, specializing the first half of the layers improves the segmentation results by an average of $1.2$ mIoU points, with only a negligible decrease in linear accuracy. In contrast, specializing the late layers yields no improvement compared to the baseline.
We attribute this to the fact that \cls and patch tokens share the representation space in the first part; once this interaction is established, further specialization has limited effect. Finally, while specializing all layers produces the highest segmentation performance, it comes with a higher memory cost and a larger drop in linear accuracy.

We further analyze how the number of specialized blocks, starting from the first, affects performance (\cref{fig:depth-seg} and \ref{fig:depth-depth}). We vary the number of blocks specialized from $0$ to $24$ (total number of blocks in ViT-L) in steps of $4$, and observe that specializing the first third of the model yields the best results, while specializing later layers degrades performance. Notably, the optimal point at one third of the model's depth coincides with a marked shift in the statistics of similarity scores shown in \cref{fig:normlayers}, which might explain the effectiveness of specializing the early layers.

\subsection{Targeted Specialization within Transformers Blocks}
\label{subsec:layers}
The previous section has shown that careful selection of transformer blocks for specialization is important for optimizing the performance.
We now explore whether further improvements can be achieved with a targeted selection of \emph{specific layers} to specialize within the transformer blocks.
In the following experiments, we specialize different layers of blocks in the first third of the model while also applying specialization to the normalization layers in all blocks.

\cref{table:layer_ablation} shows model performance on global and dense prediction tasks when specializing different layers (QKV projection, Linear and MLP, see \cref{fig:architecture}).
We observe that the performance on global task remain largely stable independently of the selected layers. In contrast, results on dense segmentation tasks get further improvements beyond what is achieved with normalization specialization alone. Interestingly, the gains do not increase monotonically with the number of specialized layers or additional parameters, as might be expected from typical scaling laws \citep{touvron2021going}. 
Specializing either or both QKV and post-attention projections consistently yields improvements.
In particular, the greatest performance gains are achieved by specializing the QKV projection, which introduces only $8\%$ additional parameters while delivering an average increase of $+1$ mIoU point over normalizations alone.
In contrast, specializing the post-attention projection does not offer further benefits, and specialization of the MLP layer either has no effect or negatively impacts performance.
Note that we could ease this 8\% memory cost overhead with Low Rank Adaptation used when specializing the QKV projection. We produce encouraging preliminary results with different ranks in \cref{subsec:lora} and leave further investigation as future work.

Our overall results show that increasing the disentanglement between \cls and patch tokens before the attention mechanism (with separated normalizations and projection) contributes to improved dense prediction performance. We hypothesize that encouraging the \cls and patch tokens to assume more distinct roles in the attention mechanism enhances their interactions, ultimately improving overall model effectiveness.
We also report results in \cref{ap:exp_without_norms} for the setting where only the QKV projections are specialized, while the normalization layers remain shared. In this configuration, performance is comparable to the baseline, indicating that the specialization of normalization layers is critical to achieve improvements, as shown in \cref{subsec:norm}.

\begin{table}[t]
\small
\centering
\caption{\textbf{Layer specialization ablation.} Performance and increase in parameter count of models trained with different layer specialization strategies, applied to the first third of the transformer blocks. In all cases, the normalization specialization described in \cref{subsec:norm} is applied in all blocks. The base model is a DINOv2 ViT-L with attention bias.}
\vspace{-6pt}
\label{table:layer_ablation}
\begin{tabular}{ccc cc cccc}
\toprule
 QKV & \multirow{2}{*}{Linear} & \multirow{2}{*}{MLP} & & & Parameter & Linear & \multirow{2}{*}{Avg. Seg.} & \multirow{2}{*}{Avg. Depth $\downarrow$} \\
 proj. & & & & & Increase (\%) & Accuracy & & \\
\midrule
 & &  & & & 0.05 & 85.1 & 65.6 & 1.178 \\ 
 & \checkmark & & & & 3 & \textbf{85.3} & 66.1 & 1.191 \\ 
 \checkmark & & &  & & 8 & \underline{85.2} & \textbf{66.6} & \underline{1.165} \\ 
 \checkmark & \checkmark & & & & 11 & 85.1 & 66.1 & 1.180 \\ 
 & & \checkmark & & &  22 & 85.1 & 65.2 & 1.189 \\ 
 & \checkmark & \checkmark & & & 25 & 85.1 & 65.9 & \textbf{1.163} \\ 
 \checkmark & & \checkmark & & & 30 & \textbf{85.3} & 65.6 & 1.185 \\ 
 \checkmark &  \checkmark & \checkmark & & & 33 & \textbf{85.3} & \underline{66.1} & 1.174 \\ 
\bottomrule
\end{tabular}
\vspace{-13pt}
\end{table}

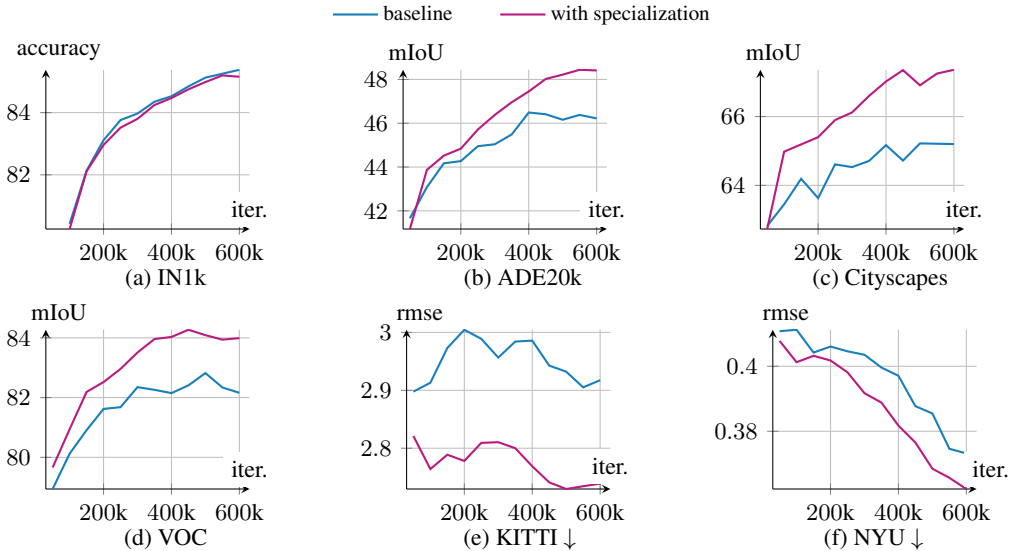
\begin{figure}[t]
\centering
\ref{mylegend}

\begin{subfigure}[b]{0.32\textwidth}
       \begin{tikzpicture}
        \begin{axis}[
             height=3.7cm,
            every node/.style={scale=1},
            xlabel={iter.},
            ylabel={accuracy},
            ymajorgrids,
            xmajorgrids,
            tick label style={font=\footnotesize},       
            label style={font=\small},                    
            ylabel near ticks,                            
            ylabel style={                                
                font=\small,
                yshift=-5pt,                             
            },
            xlabel style={
                font=\small,
                yshift=5pt, 
            },
            legend cell align={left},
            legend pos=south east,
            xmin=30, 
            xmax=630, 
            legend style={nodes={scale=0.8, transform shape}, draw=none},
            axis lines=left,
            xticklabel={$\pgfmathprintnumber{\tick}$k},
            x label style={at={(axis description cs:1.0,0.15)},anchor=north, fill=white,font=\footnotesize,},
            y label style={at={(axis description cs:0.0,1.0)},anchor=south, rotate=-90,font=\footnotesize,},
            legend to name={mylegend},
            legend style={ 
                at={(0.5,-0.15)},
                anchor=north,
                legend columns=2,
            },
            no markers,
        ]
        \addplot[scatter, no marks, draw=\baselinecolor, thick] 
            table [x=its, y={20250526_114103_large_baseline_bias_linear_acc}, col sep=comma] {figures/numbers/perfs_vs_size/perfs_vs_size.csv}; 
            \addlegendentry[color=black]{baseline$\qquad$}
        \addplot[scatter, no marks, draw=\speccolor, thick] 
            table [x=its, y={20250804_100336_double_8_split_norms_qkv_linear_acc}, col sep=comma] {figures/numbers/perfs_vs_size/perfs_vs_size.csv}; 
            \addlegendentry[color=black]{with specialization}

        \end{axis}
    \end{tikzpicture}
    \vspace{-7pt}
    \caption{IN1k}
\end{subfigure}
\hfill
\begin{subfigure}[b]{0.32\textwidth}
       \begin{tikzpicture}
        \begin{axis}[
             height=3.7cm,
            every node/.style={scale=1},
            xlabel={iter.},
            ylabel={mIoU},
            ymajorgrids,
            xmajorgrids,
            tick label style={font=\footnotesize},       
            label style={font=\small},                    
            ylabel near ticks,                            
            ylabel style={                                
                font=\small,
                yshift=-5pt,                             
            },
            xlabel style={
                font=\small,
                yshift=5pt, 
            },
            legend cell align={left},
            legend pos=south east,
            xmin=30, 
            xmax=630, 
            legend style={nodes={scale=0.8, transform shape}, draw=none},
            axis lines=left,
            xticklabel={$\pgfmathprintnumber{\tick}$k},
            x label style={at={(axis description cs:1.0,0.15)},anchor=north, fill=white,font=\footnotesize,},
            y label style={at={(axis description cs:0.0,1.0)},anchor=south, rotate=-90,font=\footnotesize,},
        ]
        \addplot[scatter, no marks, draw=\baselinecolor, thick] 
            table [x=its, y={20250526_114103_large_baseline_bias_ade20k}, col sep=comma] {figures/numbers/perfs_vs_size/perfs_vs_size.csv}; 
        \addplot[scatter, no marks, draw=\speccolor, thick] 
            table [x=its, y={20250804_100336_double_8_split_norms_qkv_ade20k}, col sep=comma] {figures/numbers/perfs_vs_size/perfs_vs_size.csv}; 

        \end{axis}
    \end{tikzpicture}
    \vspace{-7pt}
    \caption{ADE20k}
\end{subfigure}
\hfill
\begin{subfigure}[b]{0.32\textwidth}
       \begin{tikzpicture}
        \begin{axis}[
             height=3.7cm,
            every node/.style={scale=1},
            xlabel={iter.},
            ylabel={mIoU},
            ymajorgrids,
            xmajorgrids,
            tick label style={font=\footnotesize},       
            label style={font=\small},                    
            ylabel near ticks,                            
            ylabel style={                                
                font=\small,
                yshift=-5pt,                             
            },
            xlabel style={
                font=\small,
                yshift=5pt, 
            },
            legend cell align={left},
            legend pos=south east,
            xmin=30, 
            xmax=630, 
            legend style={nodes={scale=0.8, transform shape}, draw=none},
            axis lines=left,
            xticklabel={$\pgfmathprintnumber{\tick}$k},
            x label style={at={(axis description cs:1.0,0.15)},anchor=north, fill=white,font=\footnotesize,},
            y label style={at={(axis description cs:0.0,1.0)},anchor=south, rotate=-90,font=\footnotesize,},
        ]
        \addplot[scatter, no marks, draw=\baselinecolor, thick] 
            table [x=its, y={20250526_114103_large_baseline_bias_cityscapes}, col sep=comma] {figures/numbers/perfs_vs_size/perfs_vs_size.csv}; 
        \addplot[scatter, no marks, draw=\speccolor, thick] 
            table [x=its, y={20250804_100336_double_8_split_norms_qkv_cityscapes}, col sep=comma] {figures/numbers/perfs_vs_size/perfs_vs_size.csv}; 

        \end{axis}
    \end{tikzpicture}
    \vspace{-7pt}
    \caption{Cityscapes}
\end{subfigure}
\hfill
\begin{subfigure}[b]{0.32\textwidth}
       \begin{tikzpicture}
        \begin{axis}[
            every node/.style={scale=1},
            height=3.7cm,
            xlabel={iter.},
            ylabel={mIoU},
            ymajorgrids,
            xmajorgrids,
            tick label style={font=\footnotesize},       
            label style={font=\small},                    
            ylabel near ticks,                            
            ylabel style={                                
                font=\small,
                yshift=-5pt,                             
            },
            xlabel style={
                font=\small,
                yshift=5pt, 
            },
            legend cell align={left},
            legend pos=south east,
            xmin=30, 
            xmax=630, 
            legend style={nodes={scale=0.8, transform shape}, draw=none},
            axis lines=left,
            xticklabel={$\pgfmathprintnumber{\tick}$k},
            x label style={at={(axis description cs:1.0,0.15)},anchor=north, fill=white},
            y label style={at={(axis description cs:0.0,1.0)},anchor=south, rotate=-90},
        ]
        \addplot[scatter, no marks, draw=\baselinecolor, thick] 
            table [x=its, y={20250526_114103_large_baseline_bias_voc}, col sep=comma] {figures/numbers/perfs_vs_size/perfs_vs_size.csv}; 
        \addplot[scatter, no marks, draw=\speccolor, thick] 
            table [x=its, y={20250804_100336_double_8_split_norms_qkv_voc}, col sep=comma] {figures/numbers/perfs_vs_size/perfs_vs_size.csv}; 

        \end{axis}
    \end{tikzpicture}
    \vspace{-7pt}
    \caption{VOC}
\end{subfigure}
\hfill
\begin{subfigure}[b]{0.32\textwidth}
       \begin{tikzpicture}
        \begin{axis}[
             height=3.7cm,
            every node/.style={scale=1},
            xlabel={iter.},
            ylabel={rmse},
            ymajorgrids,
            xmajorgrids,
            tick label style={font=\footnotesize},       
            label style={font=\small},                    
            ylabel near ticks,                            
            ylabel style={                                
                font=\small,
                yshift=-5pt,                             
            },
            xlabel style={
                font=\small,
                yshift=5pt, 
            },
            legend cell align={left},
            legend pos=south east,
            xmin=30, 
            xmax=630, 
            legend style={nodes={scale=0.8, transform shape}, draw=none},
            axis lines=left,
            xticklabel={$\pgfmathprintnumber{\tick}$k},
            x label style={at={(axis description cs:1.0,0.17)},anchor=north, fill=white},
            y label style={at={(axis description cs:0.0,1.0)},anchor=south, rotate=-90},
        ]
        \addplot[scatter, no marks, draw=\baselinecolor, thick] 
            table [x=its, y={20250526_114103_large_baseline_bias_kitti}, col sep=comma] {figures/numbers/perfs_vs_size/perfs_vs_size.csv}; 
        \addplot[scatter, no marks, draw=\speccolor, thick] 
            table [x=its, y={20250804_100336_double_8_split_norms_qkv_kitti}, col sep=comma] {figures/numbers/perfs_vs_size/perfs_vs_size.csv}; 

        \end{axis}
    \end{tikzpicture}
    \vspace{-7pt}
    \caption{KITTI $\downarrow$}
\end{subfigure}
\hfill
\begin{subfigure}[b]{0.32\textwidth}
       \begin{tikzpicture}
        \begin{axis}[
             height=3.7cm,
            every node/.style={scale=1},
            xlabel={iter.},
            ylabel={rmse},
            ymajorgrids,
            xmajorgrids,
            tick label style={font=\footnotesize},       
            label style={font=\small},                    
            ylabel near ticks,                            
            ylabel style={                                
                font=\small,
                yshift=-5pt,                             
            },
            xlabel style={
                font=\small,
                yshift=5pt, 
            },
            legend cell align={left},
            legend pos=south east,
            xmin=30, 
            xmax=630, 
            legend style={nodes={scale=0.8, transform shape}, draw=none},
            axis lines=left,
            xticklabel={$\pgfmathprintnumber{\tick}$k},
            x label style={at={(axis description cs:1.07,0.15)},anchor=north, fill=white},
            y label style={at={(axis description cs:0.0,1.0)},anchor=south, rotate=-90},
        ]
        \addplot[scatter, no marks, draw=\baselinecolor, thick] 
            table [x=its, y={20250526_114103_large_baseline_bias_nyu}, col sep=comma] {figures/numbers/perfs_vs_size/perfs_vs_size.csv}; 
        \addplot[scatter, no marks, draw=\speccolor, thick] 
            table [x=its, y={20250804_100336_double_8_split_norms_qkv_nyu}, col sep=comma] {figures/numbers/perfs_vs_size/perfs_vs_size.csv}; 

        \end{axis}
    \end{tikzpicture}
    \vspace{-7pt}
    \caption{NYU $\downarrow$}
\end{subfigure}
    \vspace{-5pt}
\caption{
\textbf{Performances and training dynamics.} 
Performance vs. training iterations on global task ImageNet classification (IN1k) and dense tasks---segmentation (ADE20k, Cityscapes, VOC) and depth (KITTI, NYU)---with linear probing.
We compare baseline DINOv2 ViT-L with attn. bias and when specializing QKV projection in the first $1/3$ of the model and all normalizations.
}
\label{fig:perfs_losses_vitl}
    \vspace{-13pt}
\end{figure}

In \cref{fig:perfs_losses_vitl}, we compare performance dynamics of normalization and QKV projection specialization against baseline DINOv2 with attention bias.
Across all dense benchmarks---of both segmentation and depth estimation---specialization consistently enhances results. These improvements are evident from early stages of training and continue to increase over time. This trend suggests that employing specialization not only boosts performance but also contributes to more stable training dynamics.

\begin{table}
\small
\centering
\caption{\textbf{Generalizability} of the specialization on (a) DINOv2 when using different high-norm handling strategies (4 registers \citep{darcet2023vision}, attention bias \citep{an2025systematic} (`attn. bias') or none ($\varnothing$)), (b) different ViT sizes \new{on DINOv2 with attention bias framework} and (c) a supervised framework: DeiT-III. Relative difference between baseline and our specialization (`+ours') is shown in green if improvement and red otherwise.
}
\label{table:model_size}
\resizebox{\textwidth}{!}{
        \begin{tabular}{@{} l c *{8}{c@{\hspace{.2pt}}c@{\quad}} @{}}
        \toprule
        \multirow{2}{*}{Method} & \multirow{2}{*}{Size} & \multicolumn{2}{c}{Classif.} & \multicolumn{6}{c}{Segmentation} & \multicolumn{6}{c}{Depth $\downarrow$} & \multicolumn{2}{c}{\new{Detection}} \\
          \cmidrule(l){3-4} \cmidrule(l){5-10} \cmidrule(l){11-16} \cmidrule(l){17-18} 
          & & \multicolumn{2}{c}{ImNet} & \multicolumn{2}{c}{ADE} & \multicolumn{2}{c}{City} & \multicolumn{2}{c}{VOC} & \multicolumn{2}{c}{KITTI} & \multicolumn{2}{c}{NYU } & \multicolumn{2}{c}{SUN} & \multicolumn{2}{c}{\new{COCO}}\\
        \midrule
        \multicolumn{15}{l}{\textit{DINOv2 - With high-norm handling strategies}} \\
        \arrayrulecolor{gray!50}  
        \midrule
        $\varnothing$ & \multirow{2}{*}{L} & 85.3 & & 45.7 & & 64.2 & & 82.1 & & 2.868 & & 0.389 & & 0.410 & & 45.6 & \\
        +ours & & 85.3 &{\scriptsize+0.0\%}& 47.3 & \add{+3.5\%} & 66.6 & \add{+3.7\%} & 83.7 & \add{+1.9\%} & 2.787 & \add{-2.8\%} & 0.369 & \add{-5.1\%} & 0.390 & \add{-4.9\%} & 46.8 & \add{+2.6\%} \\
        \midrule
        4 registers & \multirow{2}{*}{L} & 85.3 & & 45.6 & & 64.9 & & 82.2 & & 2.893 & & 0.372 & & 0.411 & & 46.4 & \\
        +ours & & 85.3 &{\scriptsize+0.0\%}& 47.5 & \add{+4.2\%} & 65.9 & \add{+1.5\%} & 83.6 & \add{+1.7\%} & 2.906 & \sub{+0.4\%} & 0.367 & \add{-1.3\%} & 0.395 & \add{-3.9\%} & 46.8 & \add{+0.9\%} \\
        \midrule
        Attn. bias & \multirow{2}{*}{L} & 85.4 & & 46.2 & & 65.2 & & 82.2 & & 2.917 & & 0.373 & & 0.406 & & 46.0 & \\
        +ours & & 85.2 & \sub{-0.2\%} & 48.4 & \add{+4.8\%} & 67.4 & \add{+3.4\%} & 84.0 & \add{+2.2\%} & 2.739 & \add{-6.1\%} & 0.362 & \add{-2.9\%} & 0.393 & \add{-3.2\%} & 48.2 & \add{+4.8\%} \\
        \arrayrulecolor{black} 
        \midrule
        \multicolumn{15}{l}{\textit{DINOv2 - Other sizes}} \\
        \arrayrulecolor{gray!50} 
        \midrule
        Attn. bias & \multirow{2}{*}{B}  & 80.4 & & 38.3 & & 58.4 & & 76.6 & & 3.250 & & 0.462 & & 0.464 & & 39.6 &\\
        +ours & & 80.6 & \add{+0.2\%} & 38.5 & \add{+0.5\%} & 60.3 & \add{+3.3\%} & 76.5 & \sub{-0.1\%} & 3.236 & \add{-0.4\%} & 0.448 & \add{-3.0\%} & 0.470 & \sub{+1.3\%} & 39.8 & \add{+0.8\%} \\
        \midrule
        Attn. bias & \multirow{2}{*}{H} & 86.2 & & 48.1 & & 67.0 & & 83.1 & & 2.717 & & 0.359 & & 0.387 & & 49.9 \\
        +ours &  & 86.1 & \sub{-0.1\%} & 49.2 & \add{+2.3\%} & 67.1 & \add{+0.1\%} & 83.5 & \add{+0.5\%} & 2.752 & \sub{+1.3\%} & 0.344 & \add{-4.2\%} & 0.386 & \add{-0.3\%} & 49.5 & \sub{-0.8\%} \\
        \arrayrulecolor{black}  
        \midrule
        \multicolumn{15}{l}{\textit{DeiT-III}} \\
        \arrayrulecolor{gray!50}
        \midrule
        Attn. bias & \multirow{2}{*}{B} & 81.8 & & 25.4 & & 61.7 & & 48.9 & & 5.040 & & 0.747 & & 0.823 & & 32.8 \\ 
        +ours & & 81.7 & \sub{-0.1\%} & 26.3 & \add{+3.5\%} & 62.7 & \add{+1.6\%} & 50.7 & \add{+3.7\%} & 4.900 & \add{-2.8\%} & 0.732 & \add{-2.0\%} & 0.809 & \add{-1.7\%} & 32.5 & \sub{-0.9\%} \\ 
        \arrayrulecolor{black}        
        \bottomrule
        \end{tabular}
    }
\vspace{-15pt}
\end{table}

\subsection{Generalization Results}

We investigate the generalizability of our specialization approach across different variants of DINOv2, as presented in the upper part of \cref{table:model_size}.
Specifically, we train models using the DINOv2 recipe with two high-norm handler strategies—registers \citep{darcet2023vision} (“4 registers”) and attention bias \citep{an2025systematic} (“attn. bias”)—as well as without any handler. 
In all cases, we observe that specialization consistently boosts dense prediction results by up to 4.8$\%$ on ADE20k, while having a negligible effect on classification performance (no decrease greater than $0.2\%$). 
This shows that better separating the treatment of \cls and patch tokens is complementary to both high-norm handling strategies to improve dense features. 
We also investigate the specialization on DINOv2 ViT-B and ViT-H models and present results in the middle section of \cref{table:model_size}. It can be seen that our proposed specialization leads to improvements on most benchmarks, confirming its generalizability across different ViT model sizes.

We further explore the fully-supervised training setting by applying specialization to a ViT-B trained with DEIT-III \citep{touvron2022deit} strategy. We observe consistent improvements in dense prediction tasks, with gains reaching up to 3.7$\%$ on VOC. For ViT-L, specialization does not yield  benefits, likely due to the training dynamics related to the absence of a local loss to guide dense feature learning which causes the dense performance to degrade over time (we provide more details in \cref{ap:deit3_additional_results}). These results suggest that the effectiveness of the specialization may depend on training objectives, highlighting promising directions for future research.

Finally, we visualize the learned patch representations using PCA in \cref{fig:intro} for DINOv2 models trained with either registers or attention bias. In both settings, incorporating our specialization strategy produces cleaner and more semantically meaningful patch representations. Specifically, this approach reduces artifacts in textures and uniform regions, resulting in more accurate object segmentation.
More visualizations can be found in \ref{ap:qualitative}.

\begin{table}
\small
\centering
\caption{\new{\textbf{CaiT vs specialization} results. ViT-L models trained with DINOv2 framework. We compare DINOv2 baseline ($\varnothing$) with class-attention (CaiT) and specialization (ours) architectures. Results without and with attention bias (Attn. bias). Best result in bold.}
}
\label{table:class_att_vs_ours}
\resizebox{\textwidth}{!}{
        \begin{tabular}{@{} l *{8}{c} @{}}
        \toprule
        \multirow{2}{*}{Method} & Classif. & \multicolumn{3}{c}{Segmentation} & \multicolumn{3}{c}{Depth $\downarrow$} & Detection \\
          \cmidrule(l){2-2} \cmidrule(l){3-5} \cmidrule(l){6-8} \cmidrule(l){9-9} 
          & ImNet & ADE & City & VOC & KITTI & NYU  & SUN & COCO\\
        \midrule
        $\varnothing$  & \textbf{85.3} & 45.7 & 64.2 & 82.1 & 2.868 & 0.389 & 0.410 & 45.6 \\
        CaiT  & 84.0 & 43.5 & 63.7 & 80.8 & 2.968 & 0.397 & 0.421 & 43.1 \\
        Ours & \textbf{85.3} & \textbf{47.3} & \textbf{66.6} & \textbf{83.7} & \textbf{2.787} & \textbf{0.369} & \textbf{0.390} & \textbf{46.8} \\
        \midrule
        Attn. bias & \textbf{85.4} & 46.2 & 65.2 & 82.2 & 2.917 & 0.373 & 0.406 & 46.0 \\
        Attn. bias + CaiT & 85.2 & 45.3 & 65.7 & 82.0 & 2.882 & 0.374 & 0.400 & 46.8 \\ 
        Attn. bias + ours & 85.2 & \textbf{48.4} & \textbf{67.4} & \textbf{84.0} & \textbf{2.739} & \textbf{0.362} & \textbf{0.393} & \textbf{48.2} \\
        \bottomrule
        \end{tabular}
    }
\end{table}

\subsection{\new{More comparison}}

\new{
We compare our specialization approach to the class-attention mechanism from CaiT \citep{touvron2021going} in \cref{table:class_att_vs_ours}. We train both models following the DINOv2 framework, with and without attention bias. 
In CaiT architecture, patch tokens are processed through the transformer blocks, 
then the \cls is appended and updated via $2$ class-attention layers to aggregate information from the patch tokens of the last block.
We report results in \cref{table:class_att_vs_ours}, we observe that our specialization consistently outperforms the class-attention mechanism across all downstream tasks. 
}
\section{Conclusion}
\label{sec:ccl}

\vspace{-5pt}

In this work, we investigate the disentanglement of \cls and patches computations 
in Vision Transformers, focusing on their distinct roles 
and interactions. Through a comprehensive analysis, 
we demonstrate that disentangling their processing pathways and selectively 
specializing architectural layers leads to significant improvements in dense prediction tasks, 
including segmentation and depth estimation, while maintaining strong global performance.
 Our approach achieves these gains without increasing computational overhead, with minimal additional parameter cost, and generalizes across multiple ViT architectures and frameworks.
 These findings highlight the importance of tailored architectural designs and
 suggest promising directions for future research, including further exploration of
 efficient specialization strategies and applications to broader modalities and tasks.

\subsubsection*{Acknowledgments}

We thank Francisco Massa and Maximilian Seitzer for their insightful discussions
and help during the course of this work.

\bibliography{main}
\bibliographystyle{meta/iclr2026_conference}

\appendix
\section{Appendix}

\subsection{Addressing Token Interaction Anomalies}
\label{ap:anomalies}
In this work, we examine how the distinct roles of \cls{} and patch tokens affect their interactions within the model. Previous works \citep{darcet2023vision, sun2024massive} show that despite sharing computational pathways, these  different token types develop inter-dependencies that can lead to token anomalies, manifested as high-norm outliers in the patch features space. These anomalies suggest an underlying tension in how information flows between global and local representations.

\paragraph{Registers.} In order to mitigate such artifacts, observed when using different pre-training strategies \citep{oquab2023dinov2, touvron2022deit, radford2021learning}, \citet{darcet2023vision}  
propose to add learnable register tokens to the input sequence, whose roles are to replace the high-norm patches in the internal communication between patches and the \cls token. Doing so mitigates the appearance of such artifacts and boost overall results. 

\paragraph{Attention bias.}
The recent study on artifacts in Large Language Models by \citet{an2025systematic} investigates the systematic appearance of outliers which they link to the attention mechanism. 
They propose a solution consisting in adding learnable biases to the keys and values in each attention head. They analyze the equivalence of their solution compared to registers.

\begin{table}[h!]
\centering
\caption{\textbf{The impact of norm handling strategies} on DINOv2 results.}\label{wrap-tab:norm-handling}
    \vspace{-10pt}
    \begin{tabular}{l cccc}
    \toprule
    Norm. method & IN & ADE & City. 
    & NYU$\downarrow$ \\
    \toprule
    $\varnothing$ & 85.3 & 45.7 & 64.2 
    & 0.389 \\
    4 registers & 85.3 & 45.6 & 64.9 
    & \bf 0.372 \\
    attn. bias & \bf 85.4 & \bf 46.2 & \bf 65.2 
    & 0.373 \\
    \bottomrule
        \end{tabular}
\end{table} 
In our experiments, we observe that both strategies have a similar impact on high-norm artifacts and as seen in \cref{wrap-tab:norm-handling}, best overall performance is achieved when using the attention bias  (`attn. bias') strategy, with a significant improvement on segmentation benchmarks (e.g. ADE20k and Cityscapes).
To minimize confounding factors that could affect the interaction between the \cls and patch tokens, we adopt the attention bias strategy, which mitigates high-norm anomalies without introducing additional tokens.

\subsection{Effect of other Layers on \cls-Patches Similarities}
\label{ap:additional-layers-effect}

\begin{figure}[!h]
    \centering
    \begin{subfigure}[b]{0.32\textwidth}
        \includegraphics[height=3cm]{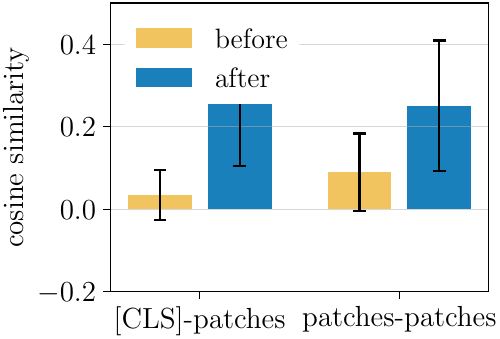}
        \caption{MLP}
    \end{subfigure}
    \begin{subfigure}[b]{0.32\textwidth}
        \includegraphics[height=3cm]{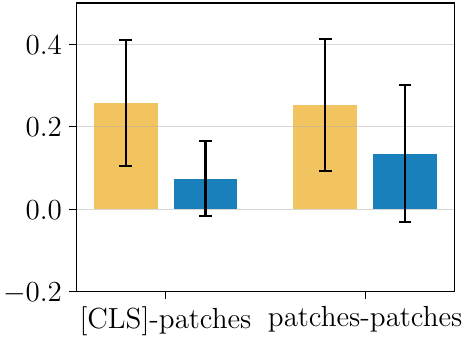}
        \caption{post-MLP LayerScale}
    \end{subfigure}
    \caption{\textbf{Effect of the MLP and post-MLP LayerScale layers on \cls-patches similarity}
    in vanilla DINOv2 pre-trained model. We show mean and standard deviation of the cosine similarity between \cls and all patches (`CLS-patches'), and between patches (`patches-patches'), before and after the considered layers.}
    \label{fig:additional-layers-effect}
\end{figure}

We report in \cref{fig:additional-layers-effect}, the effect on \cls-patches similarity of the MLP and post-MLP LayerScale layers within transformer blocks. The MLP layer, similar to the self-attention layer, increases the similarity between \cls and patches as it aligns the features. The post-MLP LayerScale, similar to other normalization layers, shows a stronger disentangling effect. 

\subsection{Training and Evaluation Details}
\label{ap:eval-details}

Throughout this work, we follow the experimental protocol of \citet{oquab2023dinov2} and 
evaluate the performance of the trained models 
on a set of global and dense task benchmarks.

\paragraph{Classification} For the global task, we perform linear probing on ImageNet classification \citep{deng2009imagenet}.
We train a linear layer with SGD for 12500 iterations, using random-resized-crop data
augmentation, and the \cls token as input for the linear layer. 
We also perform the following grid search on learning rate : $\{1.0e^{-5}, 2.0e^{-5}, 5.0e^{-5},
  0.0001,
  0.0002,
  0.0005,
  0.001,
  0.002,
  0.005,
  0.01,
0.02,
  0.05,
  0.1\}$

We then report the highest accuracy value obtained on the validation set as is common practice. 

\paragraph{Segmentation} For semantic segmentation, we use ADE20K \citep{zhou2017scene}, Cityscapes \citep{cordts2016cityscapes}, and VOC \citep{everingham2010pascal}, and report the mean Intersection over Union (mIoU) scores for each.
When we report the average performance of segmentation tasks, 
we average the scores across these $3$ datasets.
We train a linear classifier on the training
set of each benchmark for $40000$ iterations with a learning rate of $1e^{-3}$. This linear layer is applied on top of the patch output features (after the last layer normalization) of the frozen backbone, with the features further normalized using a trained batch normalization layer. 

\paragraph{Depth estimation} For depth estimation, we evaluate on KITTI \citep{geiger2013vision}, NYU Depth v2 \citep{Silberman:ECCV12}, and SUN RGB-D \citep{song2015sun}, reporting the average Root Mean Squared Error (rmse) scores. When we report the average performance of dense tasks, 
we average the scores across these $3$ datasets.
We train a linear classifier on the training
set of each benchmark for $38400$ iterations with a learning rate of $1e^{-3}$. For the input of this linear layer, we take patch and \cls output features from four evenly spaced layers of the backbone, not applying the last layer normalization. 

\new{
\paragraph{Detection} For detection, we evaluate on COCO \citep{lin2014microsoft}, reporting the average precision (AP) score. We train upon the Plain-DETR implementation \citep{lin2023detr} using the configuration provided in the official repository. More specifically, we train a RPE DETR model during $12$ epochs with a linear rate of $0.0002$ and a $1000$ steps warmup. Compared to the default configuration, and following \cite{simeoni2025dinov3} evaluation framework, we keep the ViT encoder frozen.
}

When training with DINOv2 and DeiT-III models, we follow the default configurations 
provided in the official repository, modified to add biases 
in the attention and to specialize layers or blocks.

\new{
When training with CaiT, we follow the default architecture provided in \cite{touvron2021going} and train the models using the self-supervised framework of DINOv2. We use the same hyperparameters as DINOv2 experiments. 
}

\begin{figure}[t!]
\centering
\begin{subfigure}[b]{0.32\textwidth}
    \centering
    \includegraphics[height=3cm]{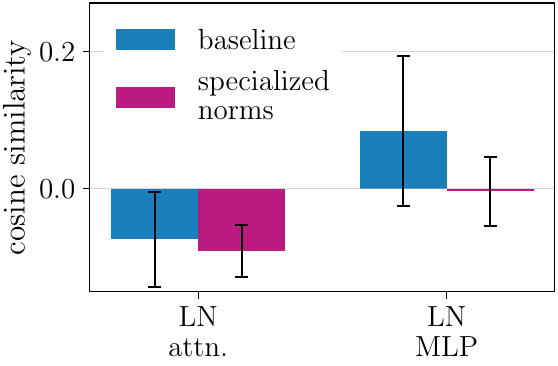}
    \caption{\cls-patch cosine similarity}
    \label{subfig:spec_impact_cls_deit}
\end{subfigure}
\begin{subfigure}[b]{0.32\textwidth}
    \centering
    \includegraphics[height=3cm]{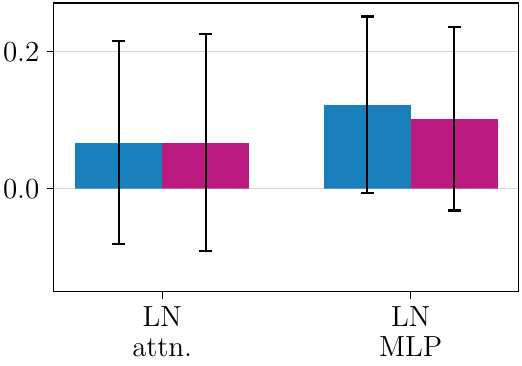}
    \caption{all-all patch cosine similarity}
    \label{subfig:spec_impact_rdm_deit}
\end{subfigure}
\caption{\textbf{Specialization of normalization layers.} Mean and standard deviation of the cosine similarity computed between (a) the \cls and all patches and (b) all patch to all patches. The average is computed over $1$k images and all model blocks. We compare post-normalization statistics between the standard architecture (`Baseline') and the model after normalization specialization (`Specialized norms'). `LN' stands for LayerNorm and `LS' for LayerScale.}
\label{fig:norm_role_deit}
\end{figure}

\subsection{Specialization of Normalization Layers in DeiT-III}
\label{ap:norm_spec_deit}

We report in \cref{fig:norm_role_deit} the impact of the specialization of the normalization layers when using DeiT-III pre-training strategy. Similar to the case of DINOv2, the average \cls-patch cosine similarity significantly reduces when employing the specialized normalization, showing the disentanglement effect. 

\subsection{LoRA Approximation}
\label{subsec:lora}

As the parameters increase can be a bottleneck for training efficiency, 
we explore the use of Low-Rank Adaptation (LoRA) \citep{hu2022lora} techniques to reduce the number of trainable parameters while maintaining performance.
Additionally, we hypothetise that \cls and patches representations share common features.
Hence we consider \cls stream as a specialization of patches stream instead of a complete different stream. Then, for a layer $f$ that we choose to specialize,
we compute the operation on the class token $x_{cls}$ as the sum of the patches layer $f_{patch}$ 
and a low-rank adaptation (LoRA) decomposition $f_{cls}^{(r)}$ of rank $r$.

\begin{figure}[!h]
    \centering
    \begin{subfigure}{0.37\textwidth}
    \includegraphics[width=\textwidth]{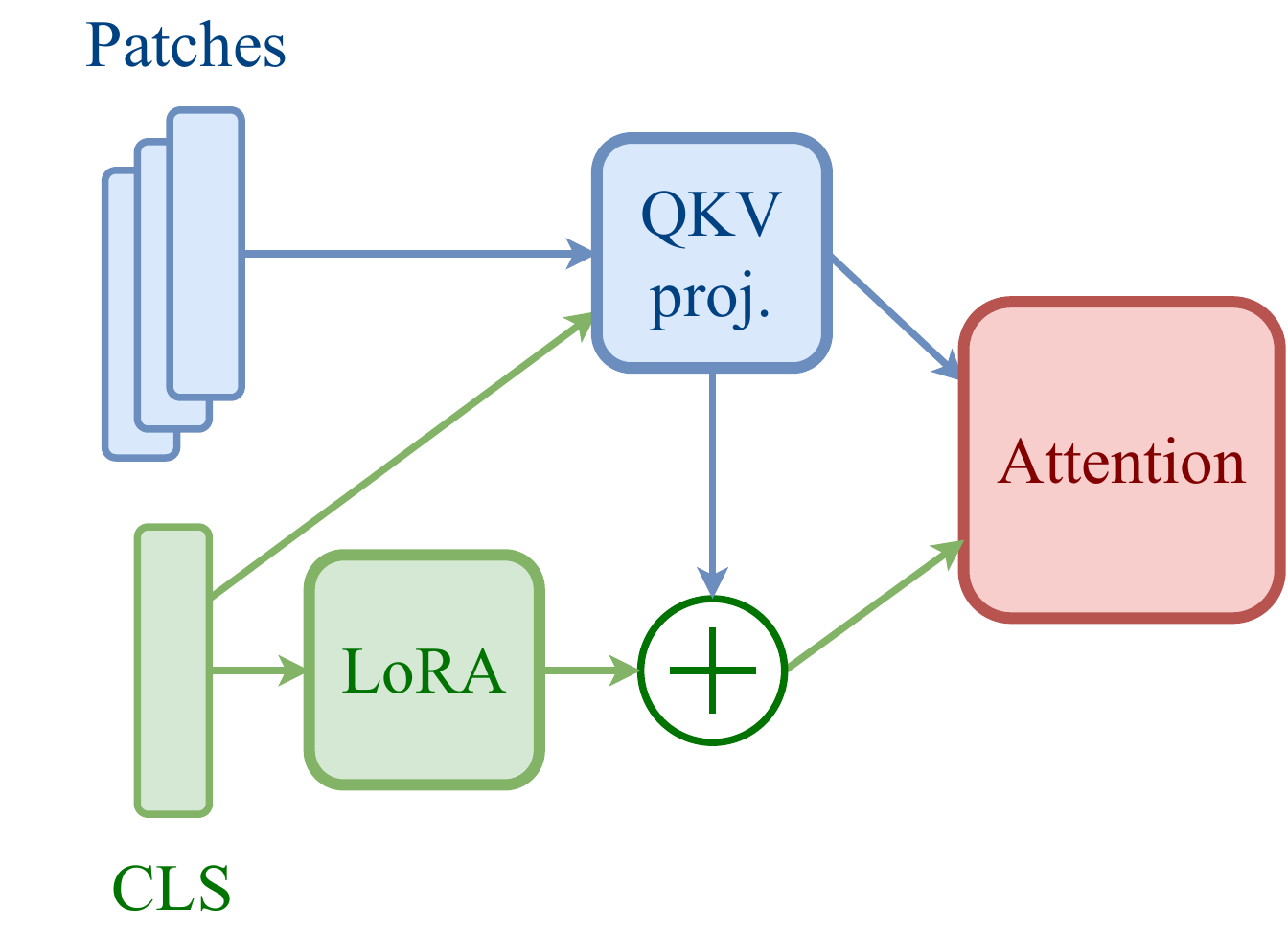}
    \caption{LoRA design}
    \label{fig:lora_design}
    \end{subfigure}
    \hfill
    \begin{subfigure}[b]{0.60\textwidth}
    \resizebox{\textwidth}{!}{
        \begin{tabular}{lcccccc}
        \toprule
         \multirow{2}{*}{Specialization}& Param. & Linear & Avg & Avg\\
          & Incr. (\%) & Acc.  &  Seg.  &  Depth $\downarrow$ \\
        \midrule
        $\varnothing$ & -- & \textbf{85.4} & 64.5 & 1.232 \\
        norms & 0.05 & 85.1 & 65.6 & \underline{1.178} \\
        +QKV & 8.3 & 85.2 & \textbf{66.6} & \textbf{1.165} \\
        \midrule
        +LoRA QKV r=16 & 0.2 & \underline{85.3} & 65.8 & 1.188 \\
        +LoRA QKV r=128 & 1.4 & 85.2 & \underline{65.9} & 1.193 \\
        \bottomrule
        \end{tabular}
    }
    \caption{Results with LoRA}
    \label{table:lora}
    \end{subfigure}
    \hfill
    \caption{\textbf{LoRA impact.} (a) Visualization of LoRA design : \cls as an approximation of patches. (b) Performance metrics and parameter increase for different LoRA configurations (rank $16$ and $128$) during first third of the model. In all cases, the normalization specialization described in \cref{subsec:norm} is applied, corresponding to 'norms' row.}
    \label{fig:lora}
    \vspace{-10pt}
\end{figure}
    
We conduct experiments in which we specialize normalization layers 
and the QKV projections with LoRA approximations of ranks $16$ and $128$
(over an embedding dimension of $1024$).
The results presented in \cref{table:lora}
shows improvements ($+0.2$ and $+0.3$ in segmentation tasks) over specializing
only the normalization layers, while adding a limited number of parameters ($+0.15\%$ and $+1.35\%$ respectively). We leave further investigations as future work.

\subsection{Normalizations Are Needed}
\label{ap:exp_without_norms}

Additionally to the specialization experiments we produced in \cref{subsec:layers}, we also conduct an experiment specializing QKV projection during the first third of
the model, \emph{but not the normalization layers}. 
We plot the results of this experiment in \cref{ap:tab:norms_importance} compared to the baseline 
and to our best model specializing normalization layers and QKV projection during third of the model. 
We observe that specializing only QKV projections brings little improvement over the baseline, e.g. +0.2 mIoU pt in average on segmentation tasks. This shows that specializing the normalization layers is crucial for best performance.

\begin{table}[t!]
\centering
\caption{\textbf{Importance of specializing norms.} Performance of for different layer specialization (Spec.) strategies applied on the first third of the transformer blocks). Normalization layers are specialized in all blocks. Baseline is a ViT-L DINOv2 with attention bias.}
\begin{tabular}{lcccccc}
\toprule
 \multirow{2}{*}{Model}& Linear & \multirow{2}{*}{Avg. Seg.} & \multirow{2}{*}{Avg. Depth $\downarrow$}\\
  & Acc. & & \\
\midrule
Baseline & \textbf{85.4} & 64.5 & 1.232 \\
Specialized norms & 85.1 & 65.6 & \underline{1.178} \\
Specialized norms \& QKV proj. & 85.2 & \textbf{66.6} & \textbf{1.165} \\
Specialized QKV proj. & 85.4 & 64.7 & 1.211 \\
\bottomrule
\end{tabular}
\label{ap:tab:norms_importance}
\end{table}

\subsection{Additional results on DeiT-III}
\label{ap:deit3_additional_results}

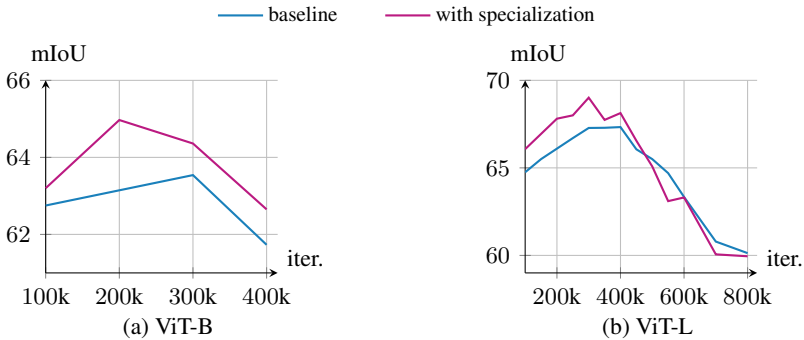
\begin{figure}[h!]
    \centering
    \ref{mylegendapp}
    
    \begin{subfigure}[b]{0.45\textwidth}
        \centering
        \begin{tikzpicture}
        \begin{axis}[
             scale=0.45,
            every node/.style={scale=1},
            xlabel={iter.},
            ylabel={mIoU},
            ymajorgrids,
            xmajorgrids,
            tick label style={font=\footnotesize},       
            label style={font=\small},                    
            ylabel near ticks,                            
            ylabel style={                                
                font=\small,
                yshift=-5pt,                             
            },
            xlabel style={
                font=\small,
                yshift=5pt, 
            },
            legend cell align={left},
            legend pos=south west,
            xmin=100, 
            xmax=415, 
            ymin=61, 
            ymax=66,
            legend style={nodes={scale=0.8, transform shape}, draw=none},
            axis lines=left,
            xticklabel={$\pgfmathprintnumber{\tick}$k},
            x label style={at={(axis description cs:1.12,0.1)},anchor=north, fill=white,font=\footnotesize,},
            y label style={at={(axis description cs:0.0,1.05)},anchor=south, rotate=-90,font=\footnotesize,},
            legend to name={mylegendapp},
            legend style={ 
                at={(0.5,-0.15)},
                anchor=north,
                legend columns=2,
            },
            no markers,
        ]
        \addplot[scatter, no marks, draw=\baselinecolor, thick] 
            coordinates {
                (100, 62.75)
                (300, 63.54)
                (400, 61.73)
            };
            \addlegendentry[color=black]{baseline$\qquad$}
            
        \addplot[scatter, no marks, draw=\speccolor, thick] 
            coordinates {
                (100, 63.2)
                (200, 64.97)
                (300, 64.36)
                (400, 62.65)
            };
            \addlegendentry[color=black]{with specialization}
            
        \end{axis}
    \end{tikzpicture}
    \vspace{-5pt}
        \caption{ViT-B}
    \end{subfigure}
    \begin{subfigure}[b]{0.45\textwidth}
        \centering
        \begin{tikzpicture}
        \begin{axis}[
             scale=0.45,
            every node/.style={scale=1},
            xlabel={iter.},
            ylabel={mIoU},
            ymajorgrids,
            xmajorgrids,
            tick label style={font=\footnotesize},       
            label style={font=\small},                    
            ylabel near ticks,                            
            ylabel style={                                
                font=\small,
                yshift=-5pt,                             
            },
            xlabel style={
                font=\small,
                yshift=5pt, 
            },
            legend cell align={left},
            legend pos=south east,
            xmin=100, 
            xmax=830, 
            ymin=59, 
            ymax=70,
            legend style={nodes={scale=0.8, transform shape}, draw=none},
            axis lines=left,
            xticklabel={$\pgfmathprintnumber{\tick}$k},
            x label style={at={(axis description cs:1.12,0.1)},anchor=north, fill=white,font=\footnotesize,},
            y label style={at={(axis description cs:0.0,1.05)},anchor=south, rotate=-90,font=\footnotesize,},
        ]
        \addplot[scatter, no marks, draw=\baselinecolor, thick] 
            coordinates {
                (100, 64.75)
                (150, 65.5)
                (250, 66.7)
                (300, 67.28)
                (350, 67.29)
                (400, 67.33)
                (450, 66.06)
                (500, 65.5)
                (550, 64.7)
                (600, 63.34)
                (700, 60.79)
                (800, 60.13)
            };
            
        \addplot[scatter, no marks, draw=\speccolor, thick] 
            coordinates {
                (100, 66.07)
                (200, 67.81)
                (250, 68.0)
                (300, 69.01)
                (350, 67.74)
                (400, 68.13)
                (450, 66.55)
                (500, 65.08)
                (550, 63.1)
                (600, 63.31)
                (700, 60.06)
                (800, 59.95)
            };
            
        \end{axis}
    \end{tikzpicture}
    \vspace{-5pt}
        \caption{ViT-L}
    \end{subfigure}
    \caption{\textbf{DeiT-III training evolution.} We visualize VOC segmentation performance (mIoU) throughout training for (a) ViT-B and (b) ViT-L pre-trained with DeiT-III (`baseline') and when adding our layer specialization.}
    \label{ap:deit_voc_vitb_vitl}
\end{figure}

We report in \cref{ap:deit_voc_vitb_vitl} the performance curves on VOC segmentation task during the training of ViT-B and ViT-L models when following DeiT-III.
We observe that the performance reaches its peak in the middle of the pre-training,
then drops significantly towards the end. We attribute this behavior to the lack of a local loss to drive dense performances. We observe a significant gain with our specialization in the first half of the training, but the gains are then diluted in the drop, particularly in the case of ViT-L.

\subsection{Other Qualitative Results}
\label{ap:qualitative}

\begin{figure}[t]
\centering
\begin{subfigure}[b]{0.32\textwidth}
\includegraphics[width=\textwidth]{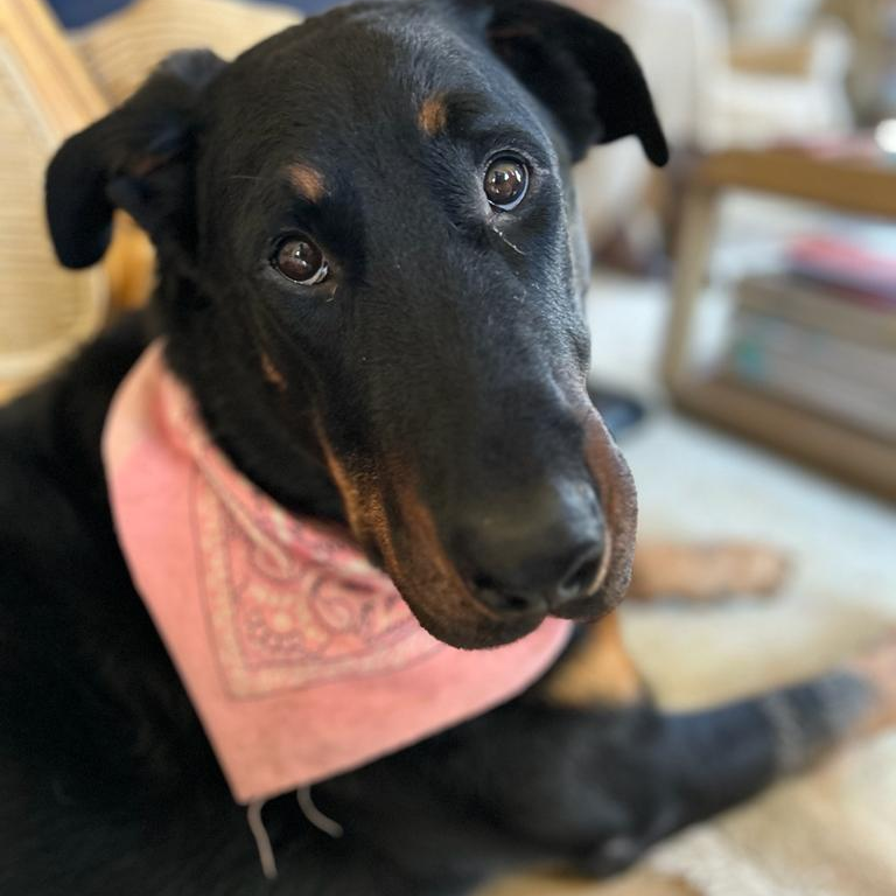}
\end{subfigure}
\hfill
\begin{subfigure}[b]{0.32\textwidth}
\includegraphics[width=\textwidth]{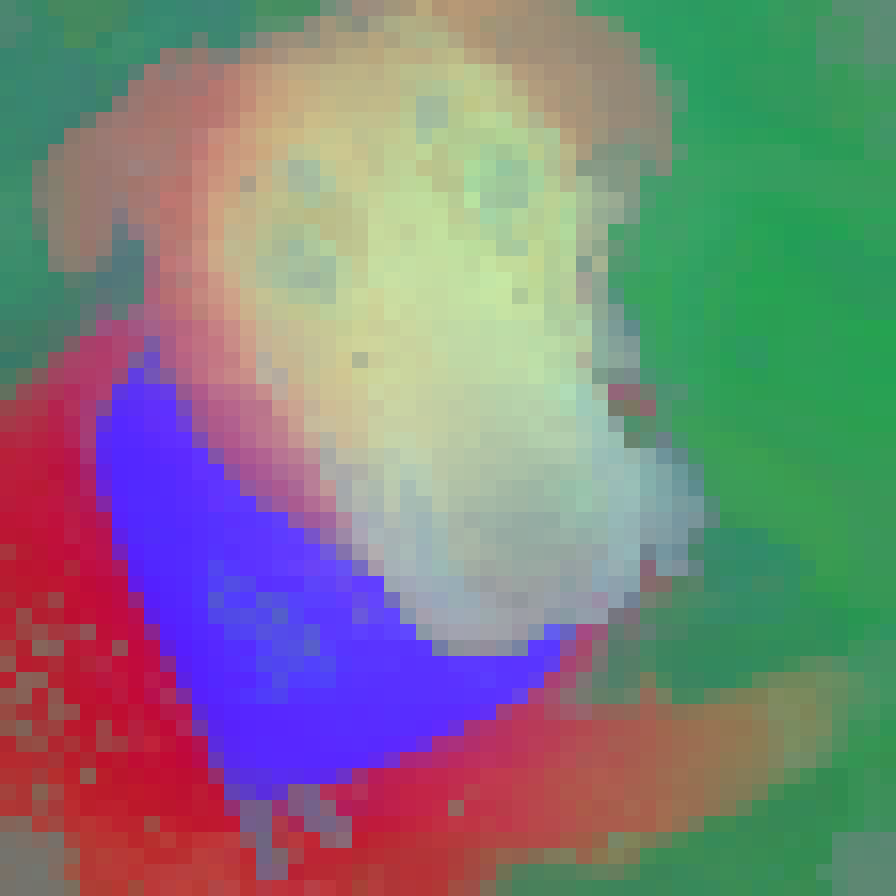}
\end{subfigure}
\hfill
\begin{subfigure}[b]{0.32\textwidth}
\includegraphics[width=\textwidth]{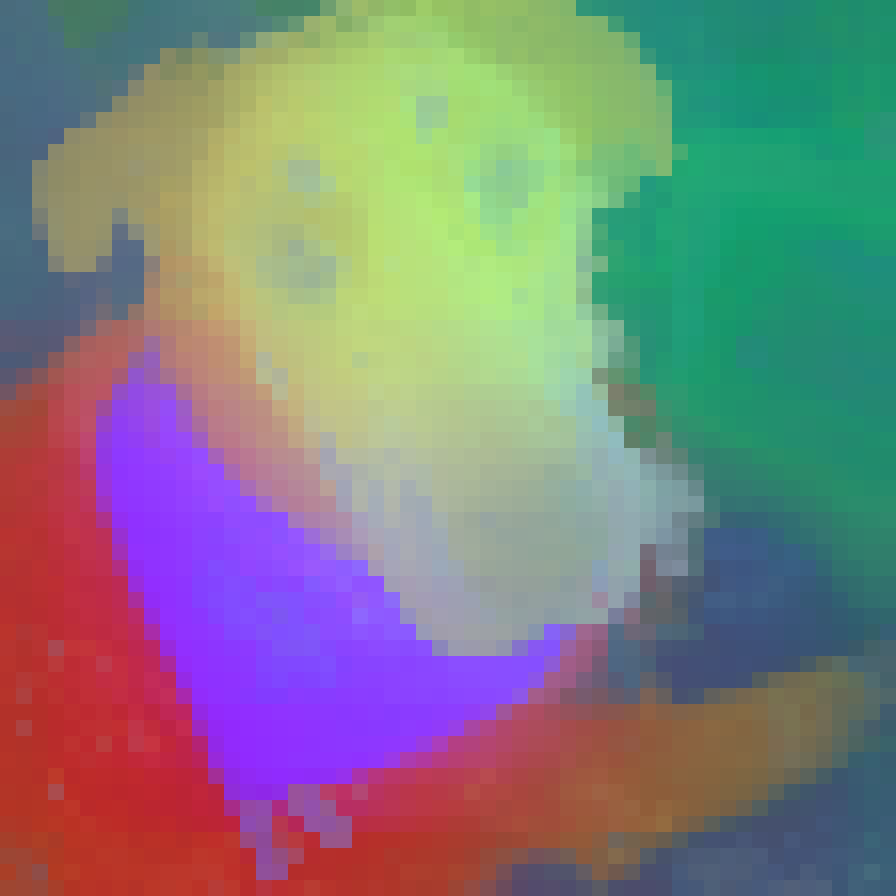}
\end{subfigure}

\begin{subfigure}[b]{0.32\textwidth}
\includegraphics[width=\textwidth]{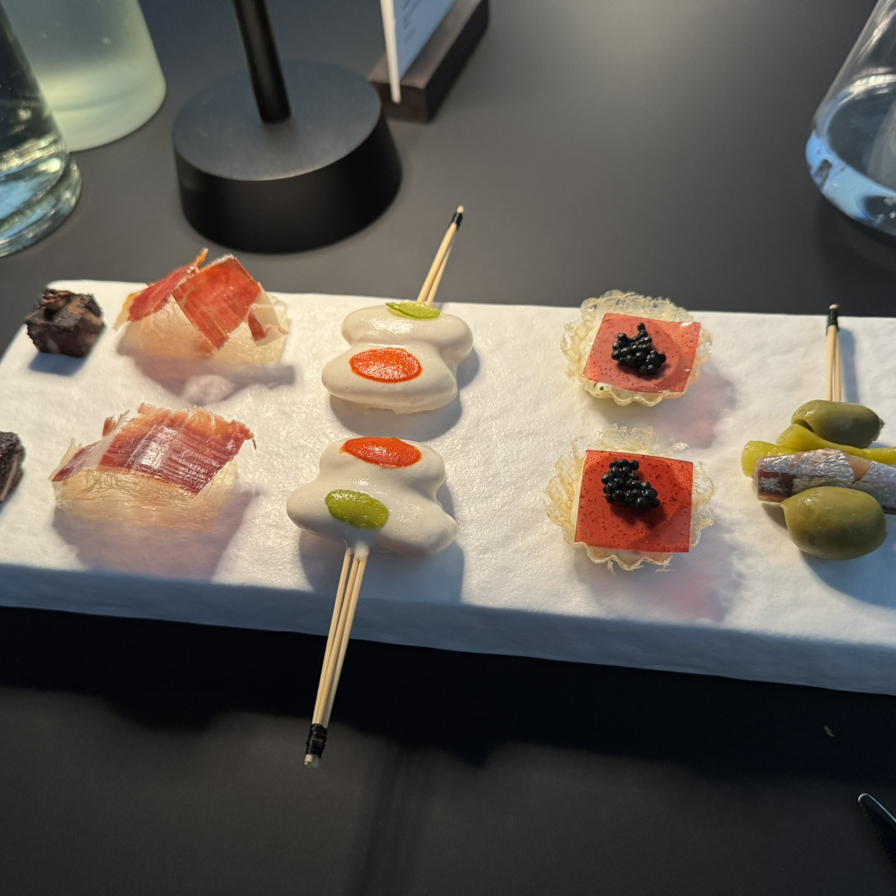}
\end{subfigure}
\hfill
\begin{subfigure}[b]{0.32\textwidth}
\includegraphics[width=\textwidth]{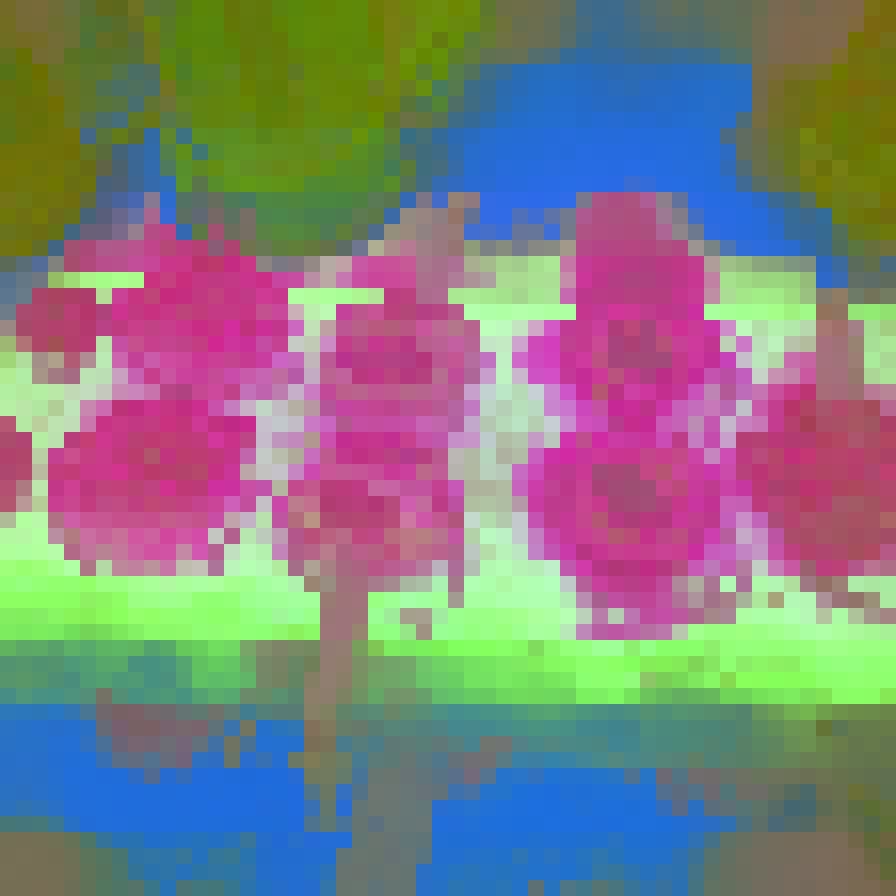}
\end{subfigure}
\hfill
\begin{subfigure}[b]{0.32\textwidth}
\includegraphics[width=\textwidth]{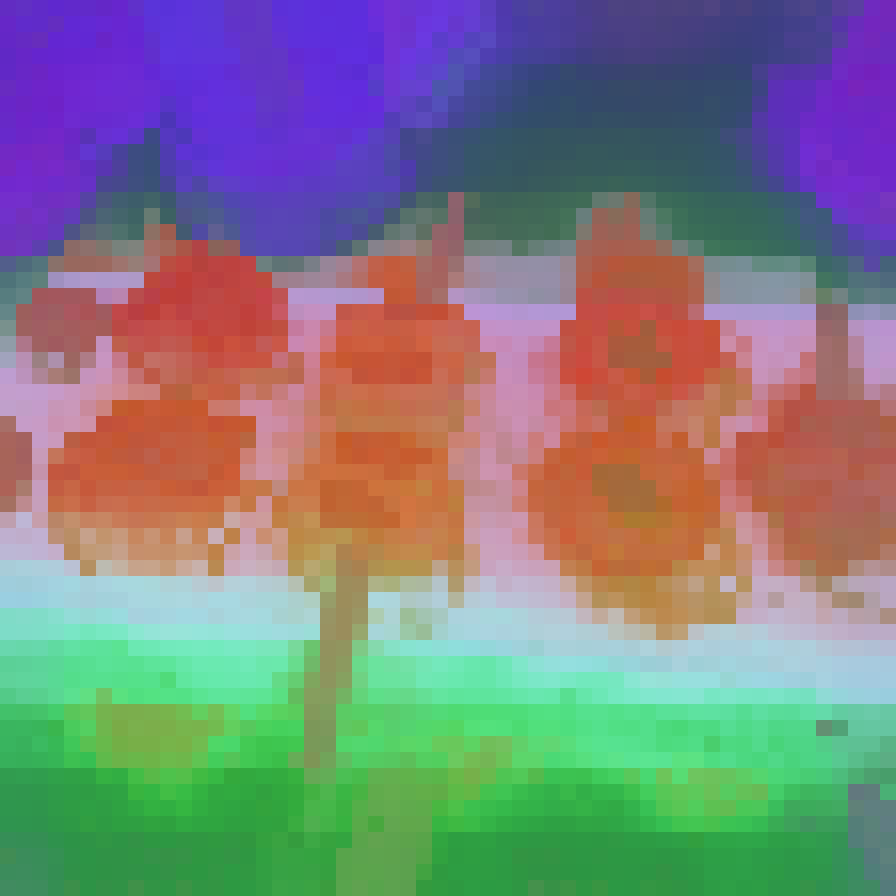}
\end{subfigure}

\begin{subfigure}[b]{0.32\textwidth}
\includegraphics[width=\textwidth]{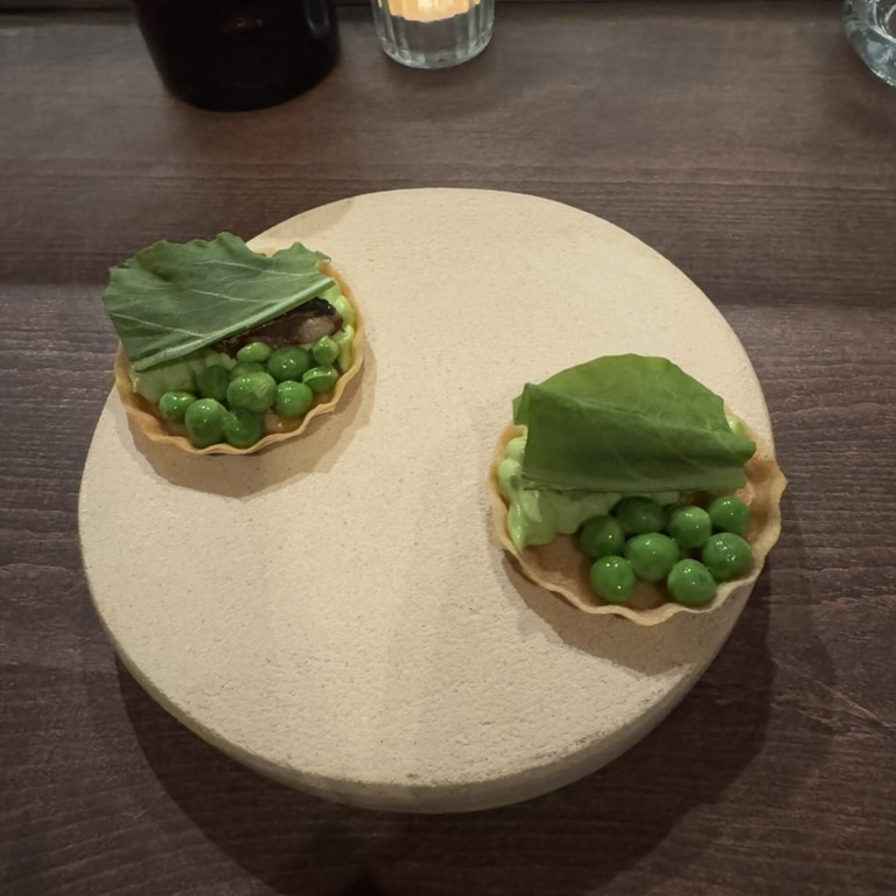}
\caption*{Original image}
\end{subfigure}
\hfill
\begin{subfigure}[b]{0.32\textwidth}
\includegraphics[width=\textwidth]{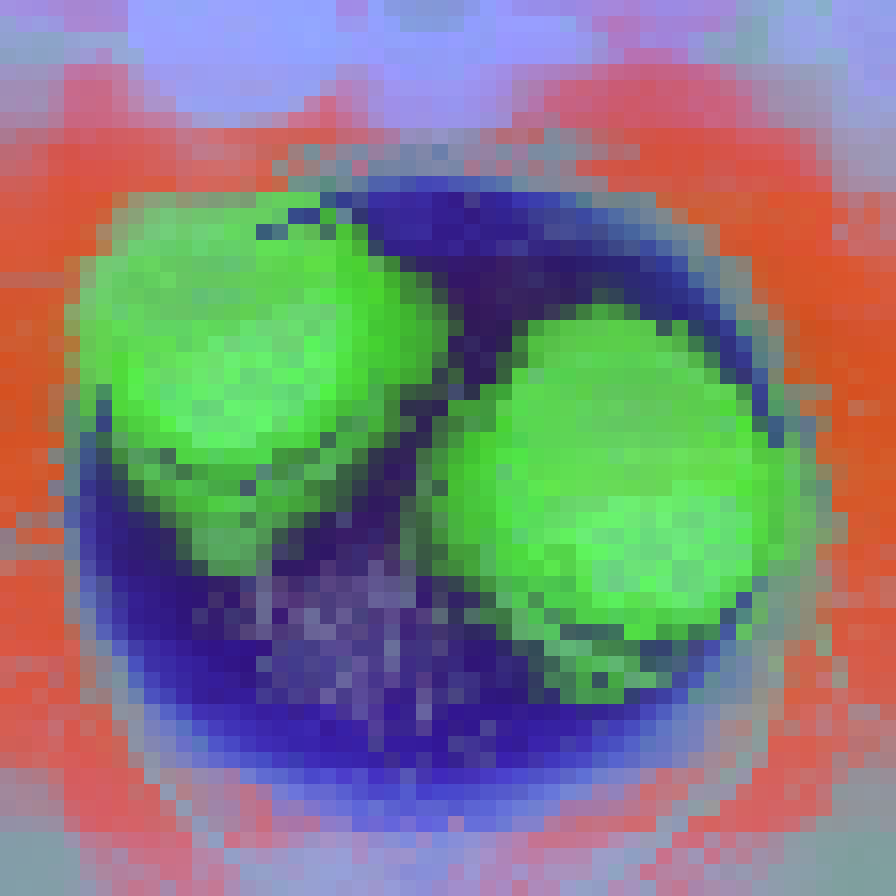}
\caption{Vanilla DINOv2}
\end{subfigure}
\hfill
\begin{subfigure}[b]{0.32\textwidth}
\includegraphics[width=\textwidth]{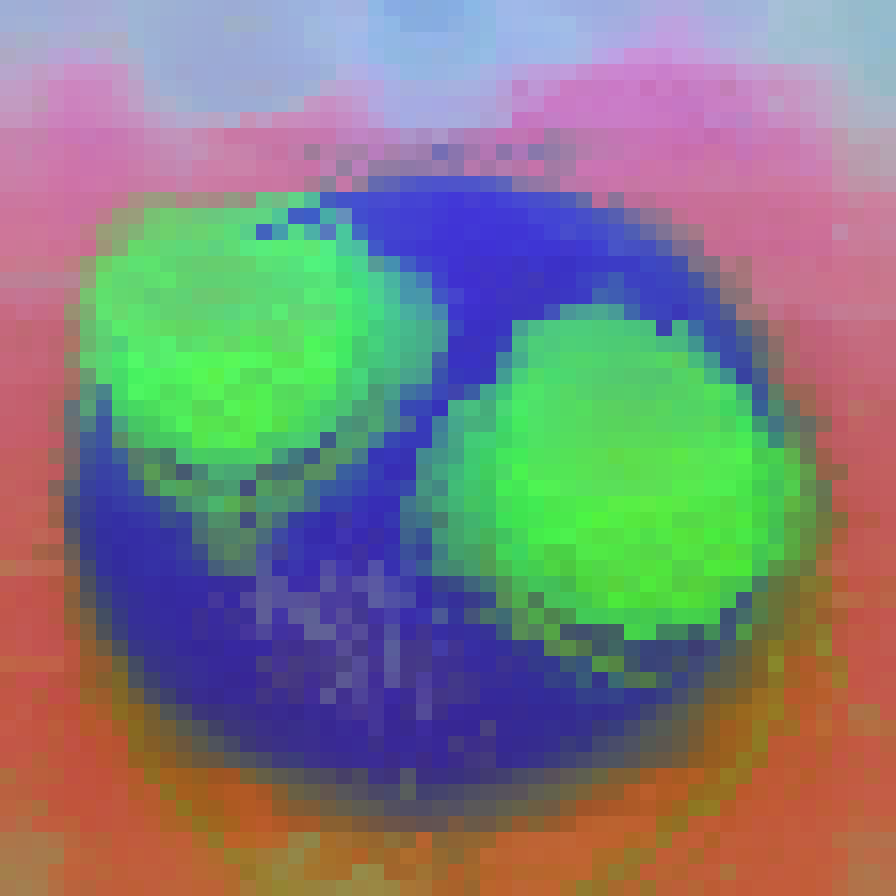}
\caption*{(a) + ours}
\end{subfigure}

\caption{First PCA components of model outputs in RGB. Specialization of normalizations and QKV projections is made during $1/3$ of the model. ViT-L with vanilla DINOv2.}
\label{fig:morequali-vanilla}
\end{figure}

\begin{figure}[t]
\centering
\begin{subfigure}[b]{0.32\textwidth}
\includegraphics[width=\textwidth]{figures/viz/pca_original_image_2.png}
\end{subfigure}
\hfill
\begin{subfigure}[b]{0.32\textwidth}
\includegraphics[width=\textwidth]{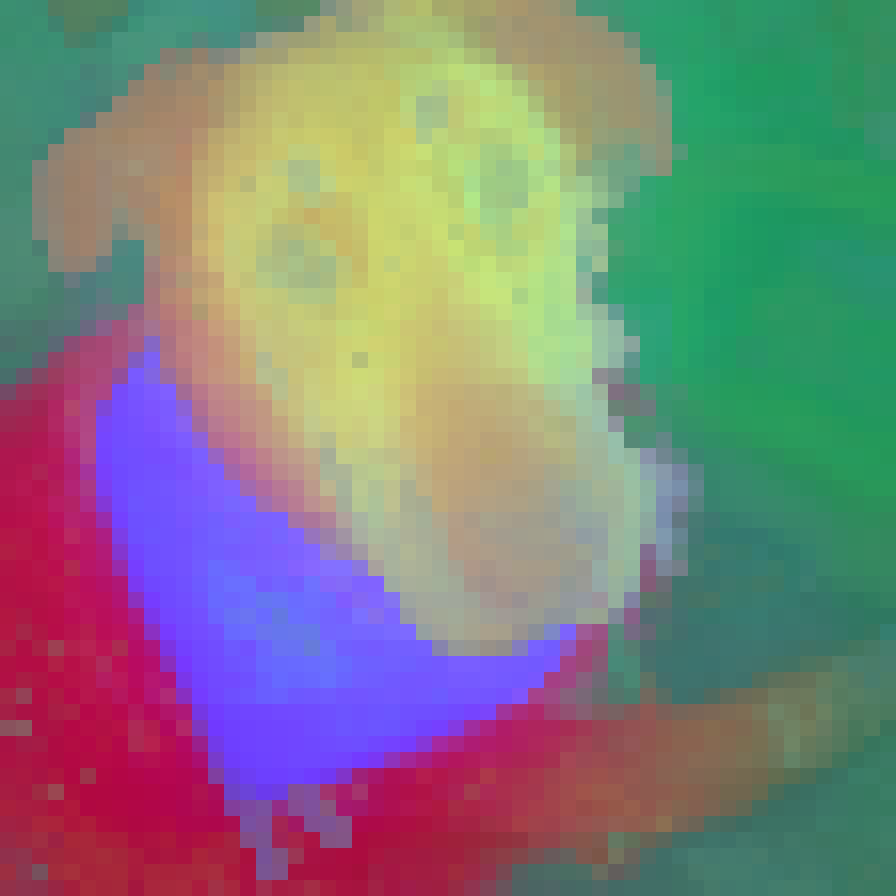}
\end{subfigure}
\hfill
\begin{subfigure}[b]{0.32\textwidth}
\includegraphics[width=\textwidth]{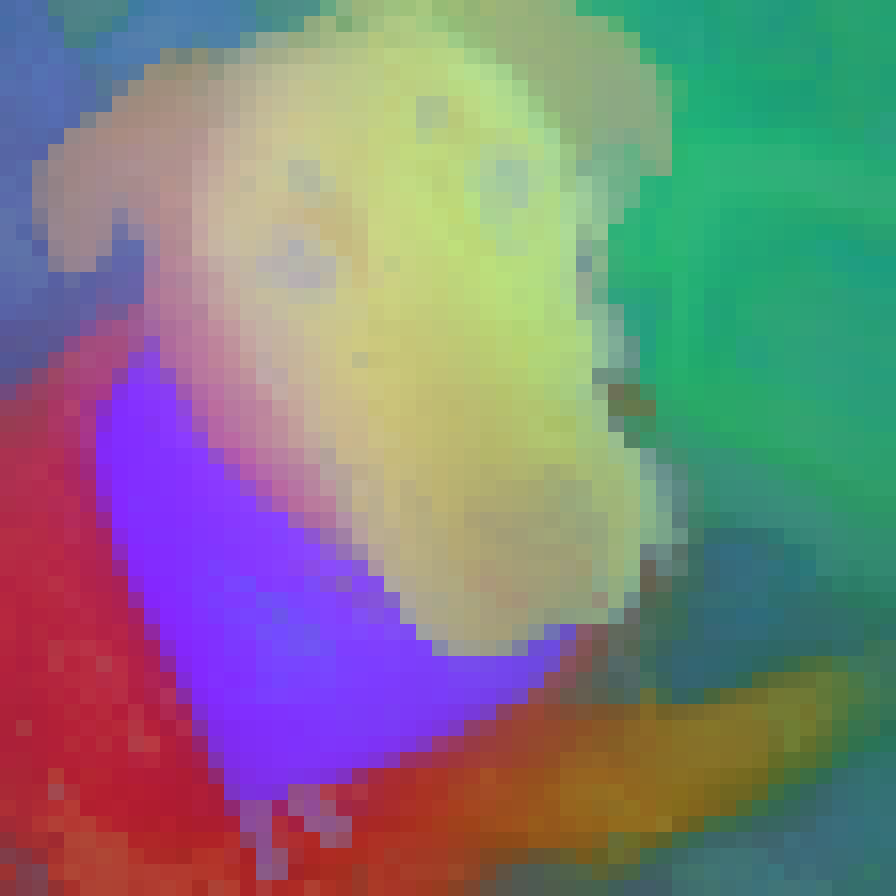}
\end{subfigure}

\begin{subfigure}[b]{0.32\textwidth}
\includegraphics[width=\textwidth]{figures/viz/pca_original_image_1.png}
\end{subfigure}
\hfill
\begin{subfigure}[b]{0.32\textwidth}
\includegraphics[width=\textwidth]{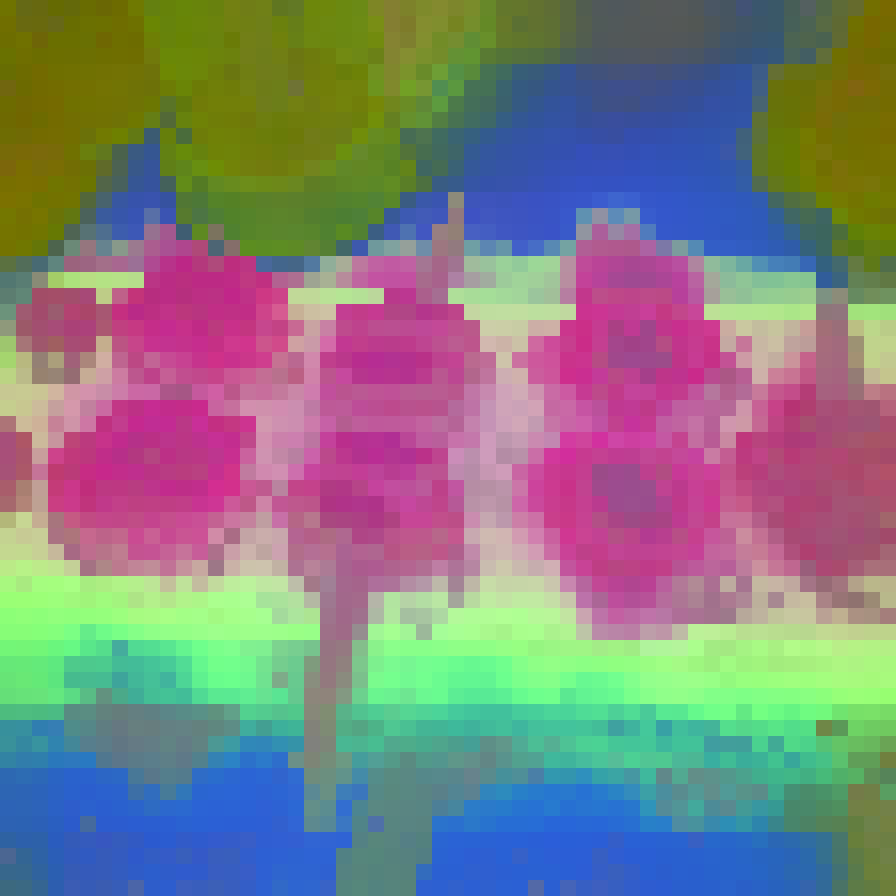}
\end{subfigure}
\hfill
\begin{subfigure}[b]{0.32\textwidth}
\includegraphics[width=\textwidth]{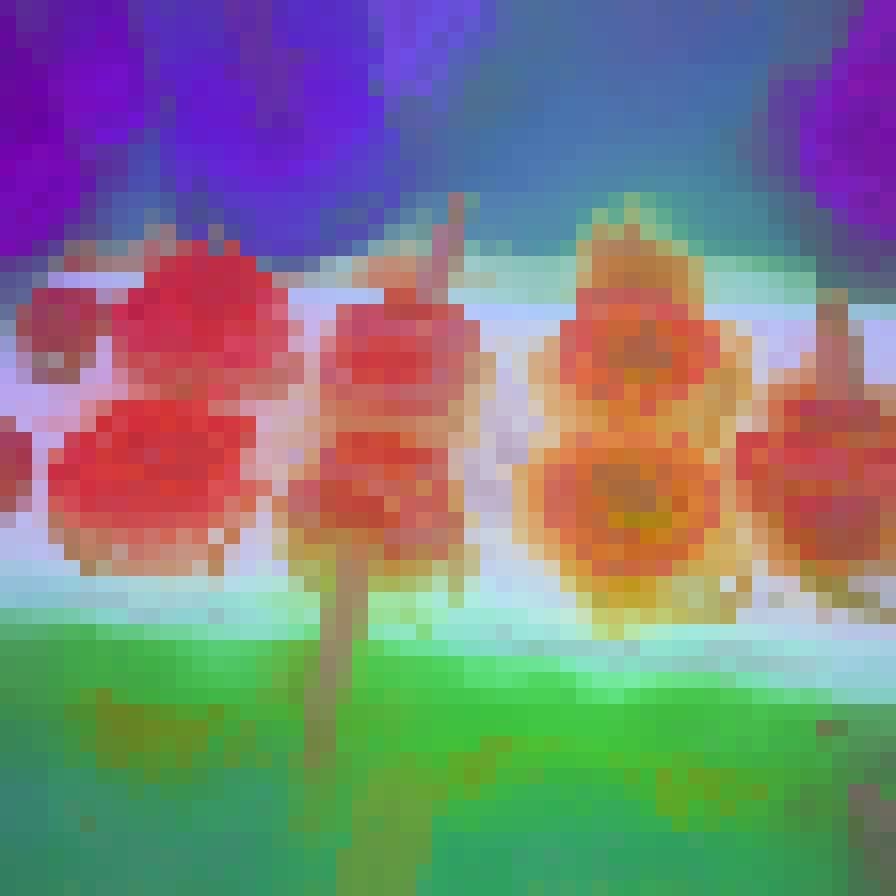}
\end{subfigure}

\begin{subfigure}[b]{0.32\textwidth}
\includegraphics[width=\textwidth]{figures/viz/pca_original_image_4.png}
\caption*{Original image}
\end{subfigure}
\hfill
\begin{subfigure}[b]{0.32\textwidth}
\includegraphics[width=\textwidth]{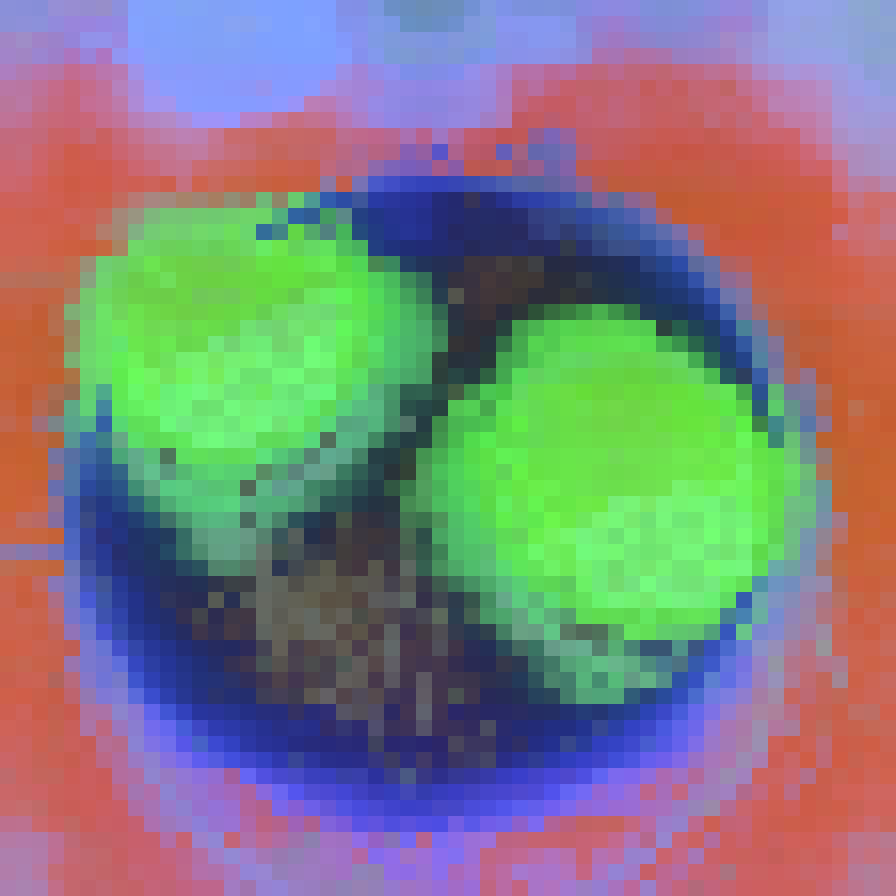}
\caption{DINOv2 w/ registers}
\end{subfigure}
\hfill
\begin{subfigure}[b]{0.32\textwidth}
\includegraphics[width=\textwidth]{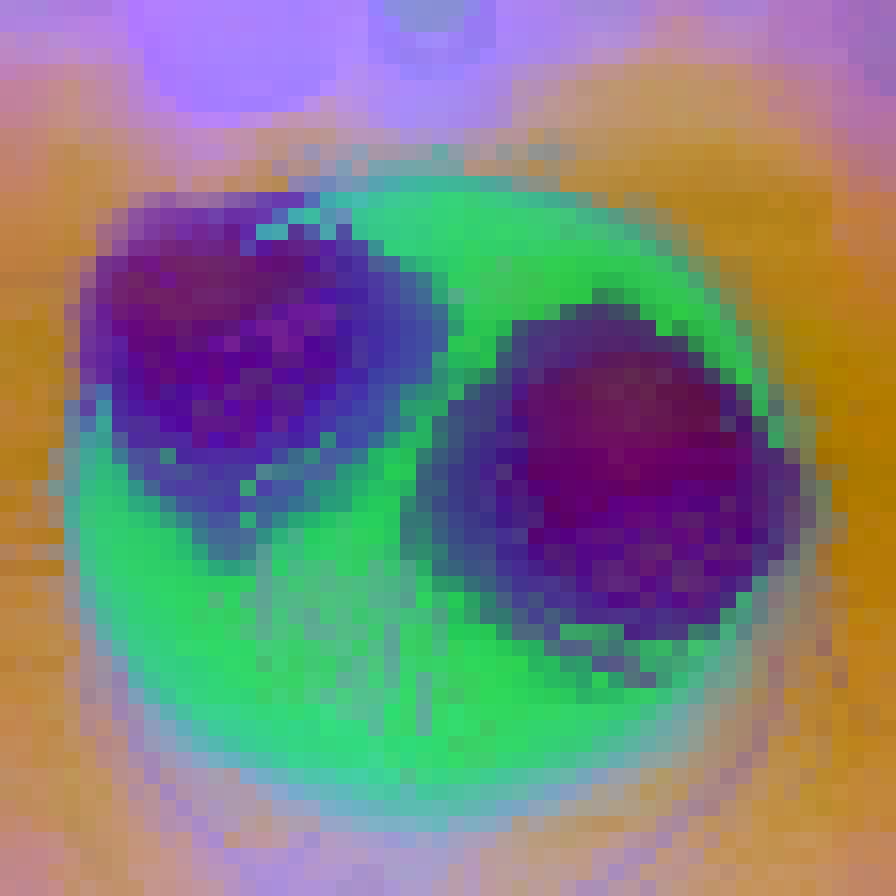}
\caption*{(a) + ours}
\end{subfigure}

\caption{First PCA components of model outputs in RGB. Specialization of normalizations and QKV projections is made during $1/3$ of the model. ViT-L DINOv2 with four registers.}
\label{fig:morequali-register}
\end{figure}

\begin{figure}[t]
\centering
\begin{subfigure}[b]{0.32\textwidth}
\includegraphics[width=\textwidth]{figures/viz/pca_original_image_2.png}
\end{subfigure}
\hfill
\begin{subfigure}[b]{0.32\textwidth}
\includegraphics[width=\textwidth]{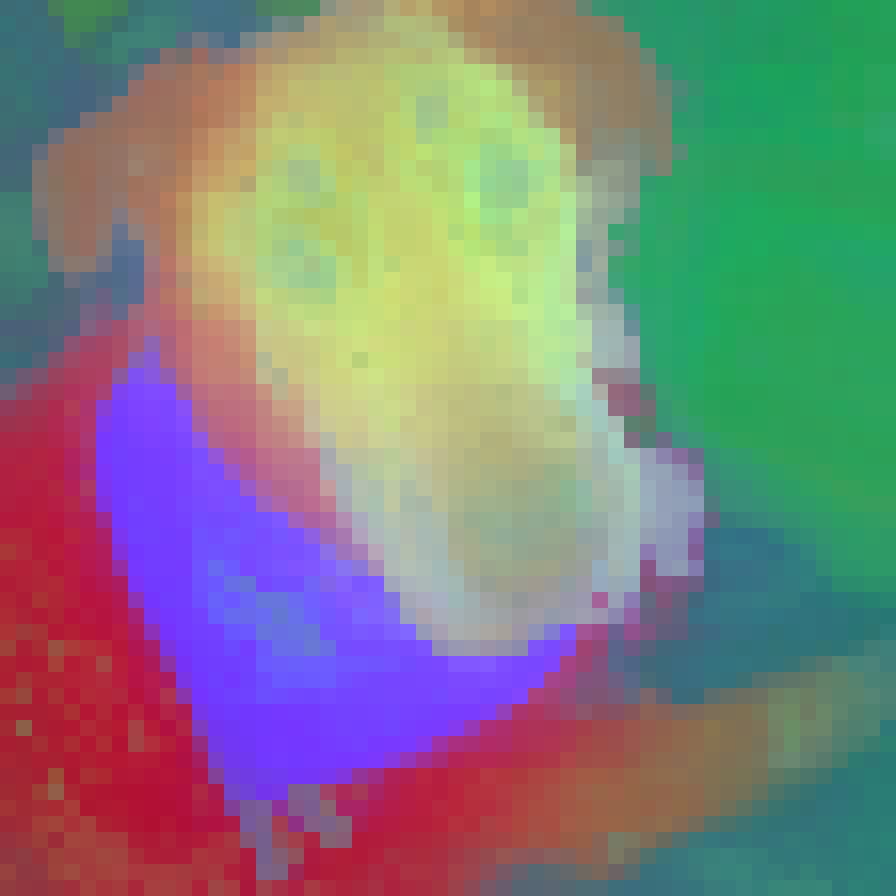}
\end{subfigure}
\hfill
\begin{subfigure}[b]{0.32\textwidth}
\includegraphics[width=\textwidth]{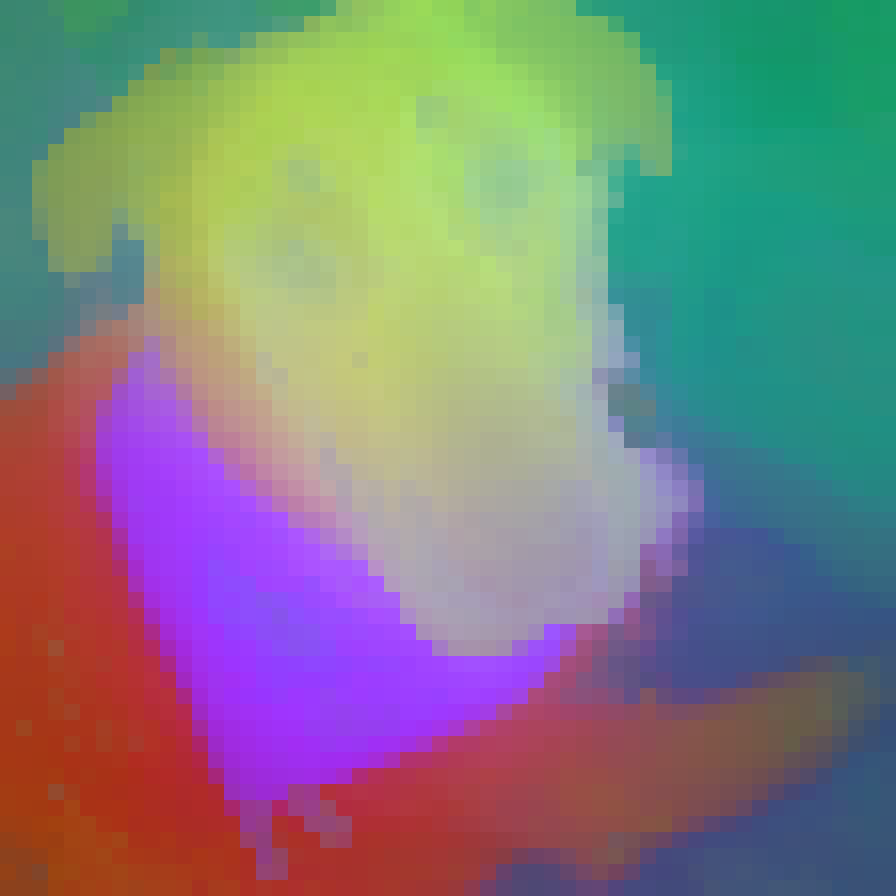}
\end{subfigure}

\begin{subfigure}[b]{0.32\textwidth}
\includegraphics[width=\textwidth]{figures/viz/pca_original_image_1.png}
\end{subfigure}
\hfill
\begin{subfigure}[b]{0.32\textwidth}
\includegraphics[width=\textwidth]{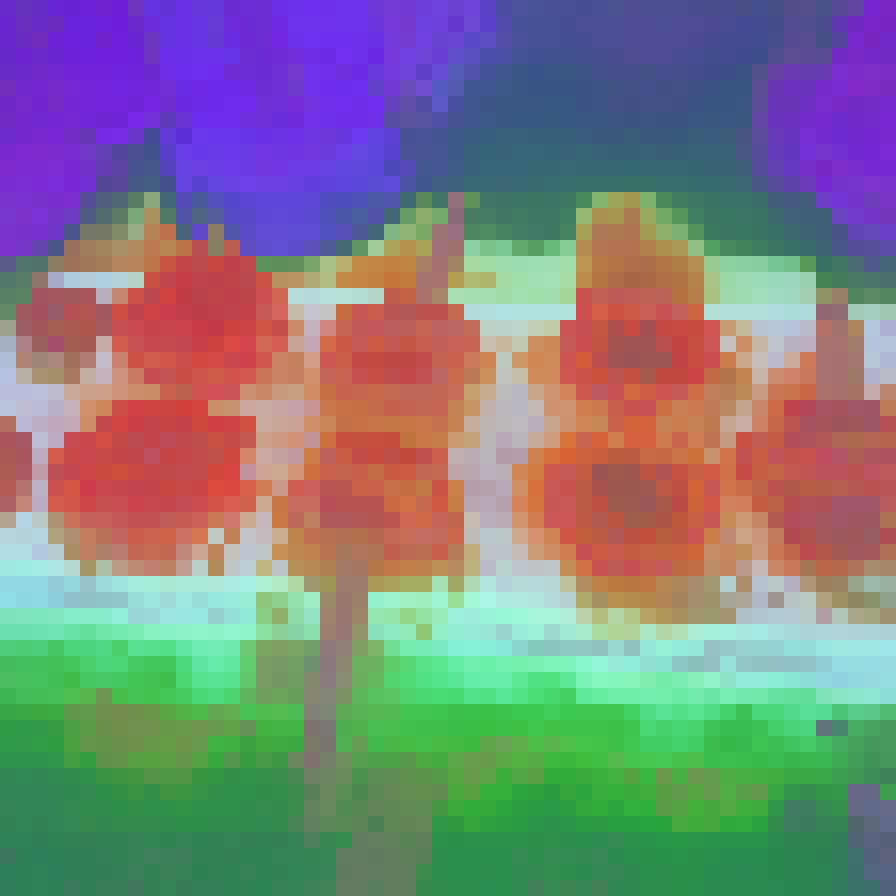}
\end{subfigure}
\hfill
\begin{subfigure}[b]{0.32\textwidth}
\includegraphics[width=\textwidth]{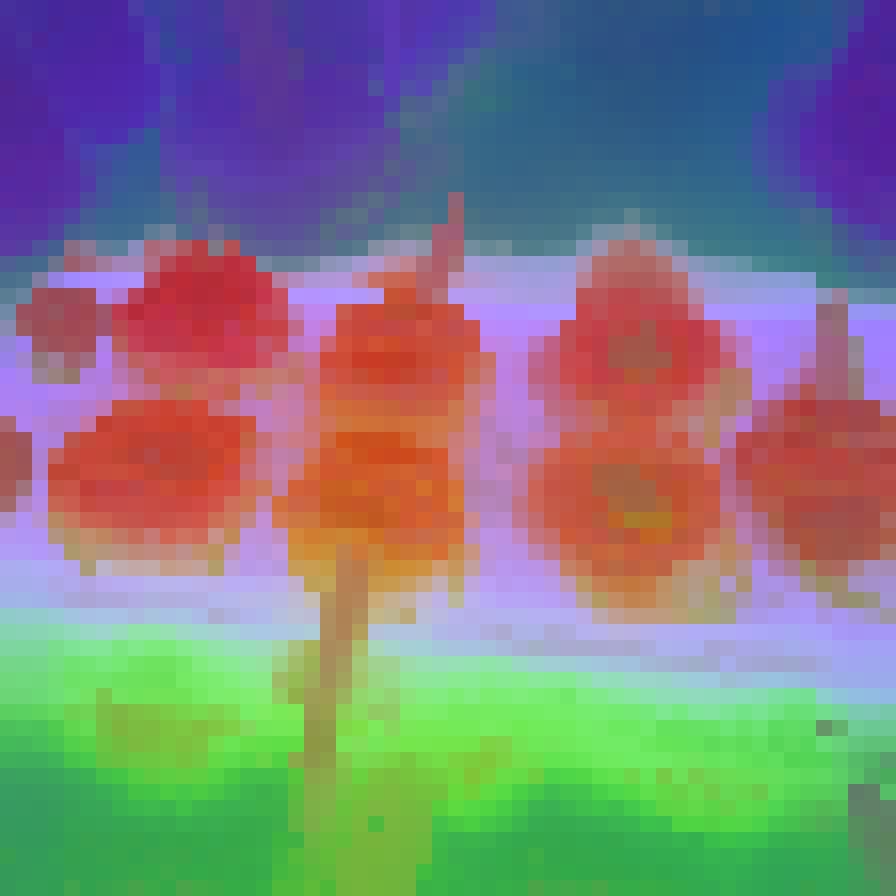}
\end{subfigure}

\begin{subfigure}[b]{0.32\textwidth}
\includegraphics[width=\textwidth]{figures/viz/pca_original_image_4.png}
\caption*{Original image}
\end{subfigure}
\hfill
\begin{subfigure}[b]{0.32\textwidth}
\includegraphics[width=\textwidth]{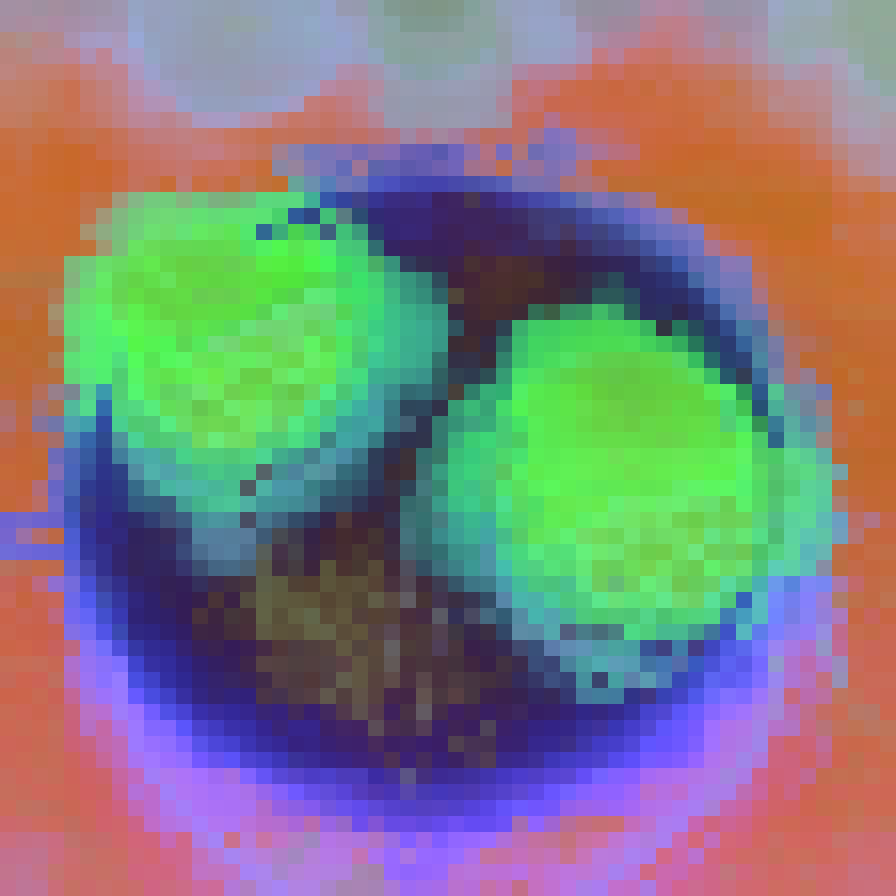}
\caption{DINOv2 w/ attn. bias}
\end{subfigure}
\hfill
\begin{subfigure}[b]{0.32\textwidth}
\includegraphics[width=\textwidth]{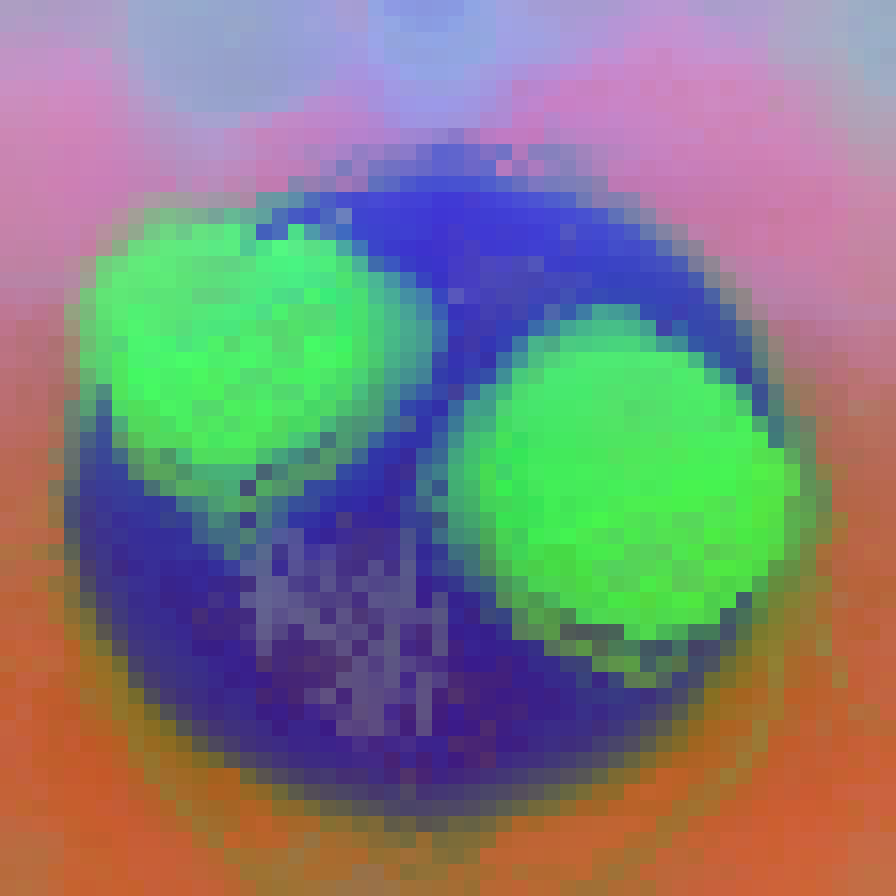}
\caption*{(a) + ours}
\end{subfigure}

\caption{First PCA components of model outputs in RGB. Specialization of normalizations and QKV projections is made during $1/3$ of the model. ViT-L DINOv2 with attention bias.}
\label{fig:morequali}
\end{figure}

We produce in \cref{fig:morequali-vanilla}, \ref{fig:morequali-register} and \ref{fig:morequali} more qualitative results when pre-training the model following DINOv2 with the vanilla architecture, four registers or attention bias and when integrating our specialization. Each figure shows the first three components, computed with the patch features, and mapped to RGB. In all cases, we observe that the specialization helps to produce more precise patch features with less artifact. For instance, we invite the reader to pay attention to the back of the dog (first row), where the artifacts visible in the original pre-training are notably reduce with our specialization. 

 \subsection{Writing details}
\label{ap:writing-details}

We have used Large Language Models (LLMs) to help write and proofread this paper. 
More specifically, they have helped to rephrase some parts of the text, propose synonyms, and check the grammar. 
We have carefully checked all the outputs of the LLMs to ensure that they are accurate and faithful to our work.

\end{document}